\documentclass[11pt,a4paper]{article}
\usepackage{authblk}
\usepackage[table,xcdraw]{xcolor}
\usepackage[hyperref]{acl2020}
\usepackage{times}
\usepackage{latexsym}
\usepackage{xurl}

\usepackage{caption}
\usepackage{subcaption}
\usepackage{graphicx,xcolor} 
\usepackage{microtype}
\usepackage{multirow}
\usepackage{enumitem}
\aclfinalcopy 
\def\aclpaperid{887} 

\title{Contributions of Transformer Attention Heads \\in  Multi- and Cross-lingual Tasks}
\author[1*]{Weicheng Ma}
\author[2*$\dagger$]{Kai Zhang}
\author[3$\dagger$]{Renze Lou}
\author[1]{Lili Wang}
\author[4]{Soroush Vosoughi}
\affil[1,4]{Department of Computer Science, Dartmouth College} 
\affil[2]{Department of Computer Science and Technology, Tsinghua University}
\affil[3]{Department of Computer Science, Zhejiang University City College}
\affil[1]{\texttt{\{first.last\}.gr@dartmouth.edu}}
\affil[2]{\texttt{drogozhang@gmail.com}}
\affil[3]{\texttt{marionojump0722@gmail.com}}
\affil[4]{\texttt{soroush.vosoughi@dartmouth.edu}}

\date{}

\begin{document}

\maketitle
\renewcommand{\thefootnote}{\fnsymbol{footnote}}
\footnotetext[1]{Equal contribution.}
\footnotetext[2]{Work done when interning at the Minds, Machines, and Society Lab at Dartmouth College.}
\renewcommand{\thefootnote}{\arabic{footnote}}
\begin{abstract}
This paper studies the relative importance of attention heads in Transformer-based models to aid their interpretability in cross-lingual and multi-lingual tasks.
Prior research has found that only a few attention heads are important in each mono-lingual Natural Language Processing (NLP) task and pruning the remaining heads leads to comparable or improved performance of the model.
However, the impact of pruning attention heads is not yet clear in cross-lingual and multi-lingual tasks.
Through extensive experiments, we show that (1) pruning a number of attention heads in a multi-lingual Transformer-based model has, in general, positive effects on its performance in cross-lingual and multi-lingual tasks and (2) the attention heads to be pruned can be ranked using gradients and identified with a few trial experiments.
Our experiments focus on sequence labeling tasks, with potential applicability on other cross-lingual and multi-lingual tasks.
For comprehensiveness, we examine two pre-trained multi-lingual models, namely multi-lingual BERT (mBERT) and XLM-R, on three tasks across 9 languages each.
We also discuss the validity of our findings and their extensibility to truly resource-scarce languages and other task settings.

\end{abstract}

\section{Introduction}

Prior research on mono-lingual Transformer-based \cite{transformer} models reveals that a subset of their attention heads makes key contributions to each task, and the models perform comparably well \cite{prune-1,prune-2} or even better \cite{prune-3} with the remaining heads pruned \footnote{We regard single-source machine translation as a mono-lingual task since the inputs to the models are mono-lingual.}.
While multi-lingual Transformer-based models, e.g. mBERT \cite{mbert-orig} and XLM-R \cite{xlm-r-orig}, are widely applied in cross-lingual and multi-lingual NLP tasks \footnote{We define a cross-lingual task as a task whose test set is in a different language from its training set. A multi-lingual task is a task whose training set is multi-lingual and the languages of its test set belong to the languages of the training set.} \cite{mbert-1,mbert-3,xlm-1}, no attempt has been made to extend the findings on the aforementioned mono-lingual research to this context.
In this paper, we explore the roles of attention heads in cross-lingual and multi-lingual tasks for two reasons.
First, better understanding and interpretability of Transformer-based models leads to efficient model designs and parameter tuning.
Second, head-pruning makes Transformer-based models more applicable to truly resource-scarce languages if it does not negatively affect model performance significantly.

The biggest challenge we face when studying the roles of attention heads in cross-lingual and multi-lingual tasks is locating the heads to prune.
Existing research has shown that each attention head is specialized to extract a collection of linguistic features, e.g., the middle layers of BERT mainly extract syntactic features \cite{probing-1,probing-2} and the fourth head on the fifth layer of BERT greatly contributes to the coreference resolution task \cite{probing-3}.
Thus, we hypothesize that important feature extractors for a task should be shared across languages and the remaining heads can be pruned.
We evaluate two approaches used to rank attention heads, the first of which is layer-wise relevance propagation (LRP, \citet{lrp-orig}).
\citet{prune-1} interpreted the adaptation of LRP in Transformer-based models on machine translation.
Motivated by \citet{gradient-orig} and \citet{gradient-1}, we design a second ranking method based on gradients since the gradients on each attention head reflect its contribution to the predictions.

We study the effects of pruning attention heads on three sequence labeling tasks, namely part-of-speech tagging (POS), named entity recognition (NER), and slot filling (SF).
We focus on sequence labeling tasks since they are more difficult to annotate than document- or sentence-level classification datasets and require more treatment in cross-lingual and multi-lingual research.
We choose POS and NER datasets in 9 languages, where English (EN), Chinese (ZH), and Arabic (AR) are candidate source languages.
The MultiAtis++ corpus \cite{mbert-2} is used in the SF evaluations with EN as the source language.
We do not include syntactic chunking and semantic role labeling tasks due to lack of availability of manually written and annotated corpora.
In these experiments, we rank attention heads based only on the source language(s) to ensure the extensibility of the learned knowledge to cross-lingual tasks and resource-poor languages.

In our preliminary experiments comparing the gradient-based method and LRP, the average F1 score improvements on NER with mBERT are 0.69 (cross-lingual) and 0.24 (multi-lingual) for LRP and 0.81 (cross-lingual) and 0.31 (multi-lingual) for the gradient-based method, though both methods rank attention heads similarly.

Thus we choose the gradient-based method to rank attention heads in all our experiments.

Our evaluations confirm that only a subset of attention heads in each Transformer-based model makes key contributions to each cross-lingual or multi-lingual task and that these heads are shared across languages.

Performance of models generally drop when the highest-ranked or randomly selected heads are pruned, validating the head rankings generated by our gradient-based method.
We also observe performance improvements on tasks with multiple source languages by pruning attention heads.
Our findings potentially apply to truly resource-scarce languages since we show that the models perform better with attention heads pruned when fewer training instances are available in the target languages.

The contributions of this paper are three-fold:
\begin{itemize}[leftmargin=*,topsep=0pt]
\setlength{\itemsep}{0cm}
\setlength{\parskip}{0cm}
    \item We explore the roles of attention heads in multi-lingual Transformer-based models and find that pruning certain heads leads to comparable or better performance in cross-lingual and multi-lingual sequence labeling tasks.
    \item We adapt a gradient-based method to locate attention heads that can be pruned without exhaustive experiments on all possible combinations.
    \item We show the correctness, robustness, and extensibility of the findings and our head ranking method under a wide range of settings through comprehensive experiments.
\end{itemize}

\section{Datasets}
\begin{table}[]
\begin{tabular}{|l|l|r|r|}
\hline
\multicolumn{1}{|l|}{\multirow{2}{*}{LC}} & \multicolumn{1}{l|}{\multirow{2}{*}{Language Family}} & \multicolumn{2}{c|}{Training Size}                  \\ \cline{3-4} 
\multicolumn{1}{|c|}{}                      & \multicolumn{1}{c|}{}                     & \multicolumn{1}{c|}{POS} & \multicolumn{1}{c|}{NER} \\ \hline
EN                                          & IE, Germanic                              & 12,543                    & 14,987                    \\ \hline
DE                                          & IE, Germanic                              & 13,814                   &  12,705                        \\ \hline
NL                                          & IE, Germanic                              & 12,264                   &  15,806                        \\ \hline
AR                                          & Afro-Asiatic, Semitic                     & 6,075                   &  1,329                        \\ \hline
HE                                          & Afro-Asiatic, Semitic                     & 5,241                   &  2,785                        \\ \hline
ZH                                          & Sino-Tibetan                              & 3,997                    & 20,905                         \\ \hline
JA                                          & Japanese                               & 7,027                    &  800                        \\ \hline
UR                                          & IE, Indic                                 & 4,043                    & 289,741                         \\ \hline
FA                                          & IE, Iranian                                    & 4,798                    & 18,463                         \\ \hline
\end{tabular}
\caption{Details of POS and NER datasets in our experiments. LC refers to language code. Training size denotes the number of training instances.}
\label{tbl:dataset-description}
\end{table}

We use human-written and manually annotated datasets in experiments to avoid noise from machine translation and automatic label projection.

We choose POS and NER datasets in 9 languages, namely EN, ZH, AR, Hebrew (HE), Japanese (JA), Persian (FA), German (DE), Dutch (NL), and Urdu (UR).
As Table \ref{tbl:dataset-description} shows, these languages fall in diverse language families and the datasets are very different in size.
EN, ZH, and AR are used as candidate source languages since they are resource-rich in many NLP tasks.
Our POS datasets are all from Universal Dependencies (UD) v2.7 \footnote{\url{http://universaldependencies.org/}}.
These datasets are labeled with a common label set containing 17 POS tags.

For NER, we use NL, EN, and DE datasets from CoNLL-2002 and 2003 challenges \cite{conll-2002,conll-2003}.
Additionally, we use the People's Daily dataset \footnote{\url{http://github.com/OYE93/Chinese-NLP-Corpus/tree/master/NER/People's Daily}}, iob2corpus \footnote{\url{http://github.com/Hironsan/IOB2Corpus}}, AQMAR \cite{ar-ner-dataset}, ArmanPerosNERCorpus \cite{fa-ner-dataset}, MK-PUCIT \cite{ur-ner-dataset}, and a news-based NER dataset \cite{he-ner-dataset} for the languages CN, JA, AR, FA, UR, and HE, respectively.
Since the NER datasets are individually constructed in each language, their label sets do not fully agree.
As there are four NE types (PER, ORG, LOC, MISC) in the three source-language datasets,
we merge other NE types into the MISC class to allow cross-lingual evaluations.

We evaluate SF models on MultiAtis++ with EN as the source language and Spanish (ES), Portuguese (PT), DE, French (FR), ZH, JA, Hindi (HI), and Turkish (TR) as target languages.
There are 71 slot types in the TR dataset, 75 in the HI dataset, and 84 in the other datasets.
We do not use the intent labels in our evaluations since we study only sequence labeling tasks.
Thus our results are not directly comparable with \citet{mbert-2}.

\section{Methodology}

Here, we introduce the gradient-based method we use in the experiments to rank the attention heads.
\citet{gradient-orig} claim that gradients measure the importance of features to predictions.
Since each head functions similarly as a standalone feature extractor in a Transformer-based model,
we use gradients to approximate the importance of the feature set extracted by each head and rank the heads accordingly.
\citet{prune-2} determine importance of heads with accumulated gradients at each head in a training epoch.
Different from their approach, we fine-tune the model on the training set and rank the heads using gradients on the development set to ensure that the head importance rankings are not significantly correlated with the training instances in one source language.

Specifically, our method generates head rankings for each language in three steps:\\
(1) We fine-tune a Transformer-based model on a mono-lingual task for three epochs.\\
(2) We re-run the fine-tuned model on the development partition of the dataset with back-propagation but not parameter updates to obtain gradients.\\
(3) We sum up the absolute gradients on each head, layer-wise normalize the accumulated gradients, and scale them into the range [0, 1] globally.\\

\begin{figure}[t]
\centering
\begin{subfigure}{.49\linewidth}
    \centering
    \includegraphics[height=1\linewidth]{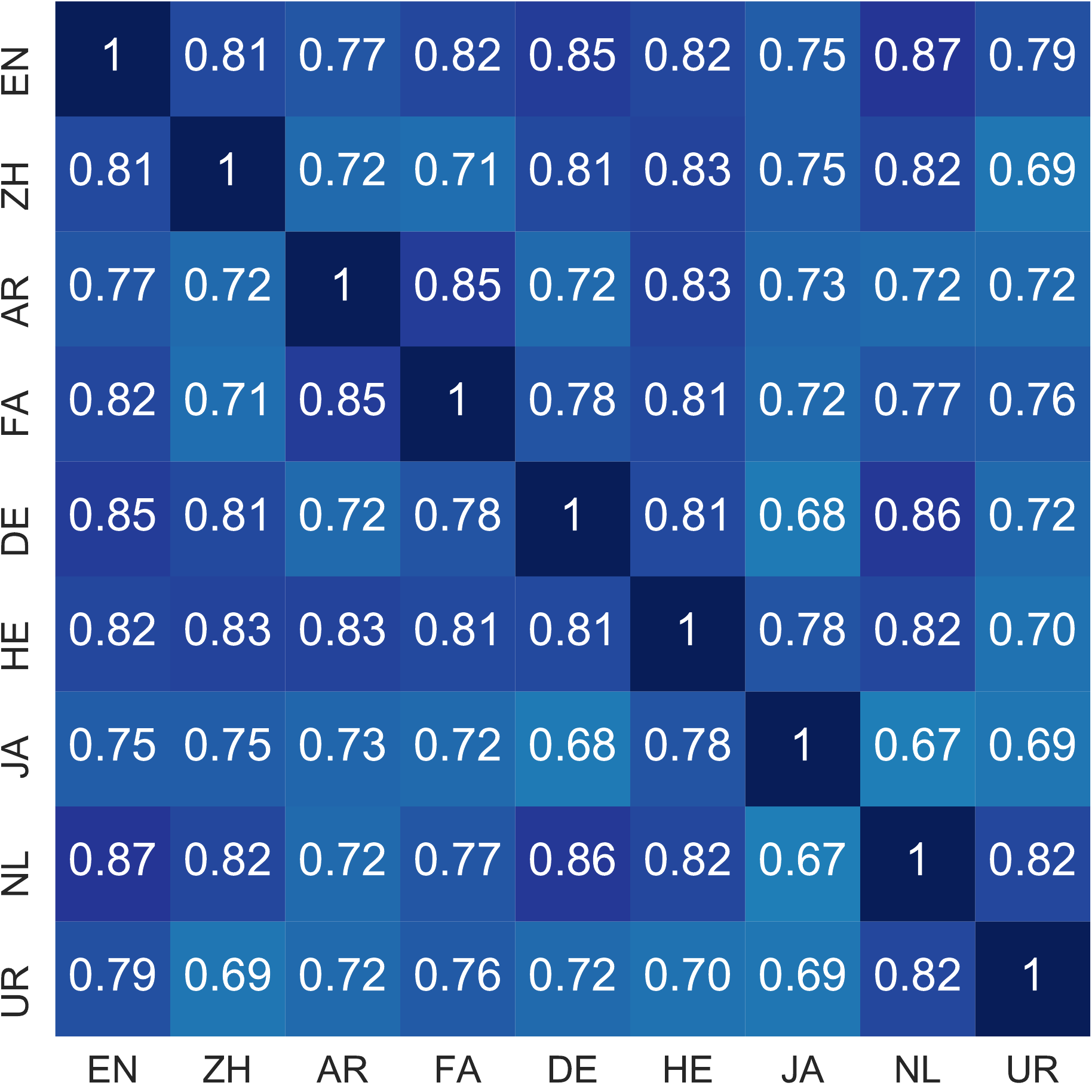}
    \caption{POS-mBERT}
\end{subfigure}
    \hfill
\begin{subfigure}{.49\linewidth}
    \centering
    \includegraphics[height=1\linewidth]{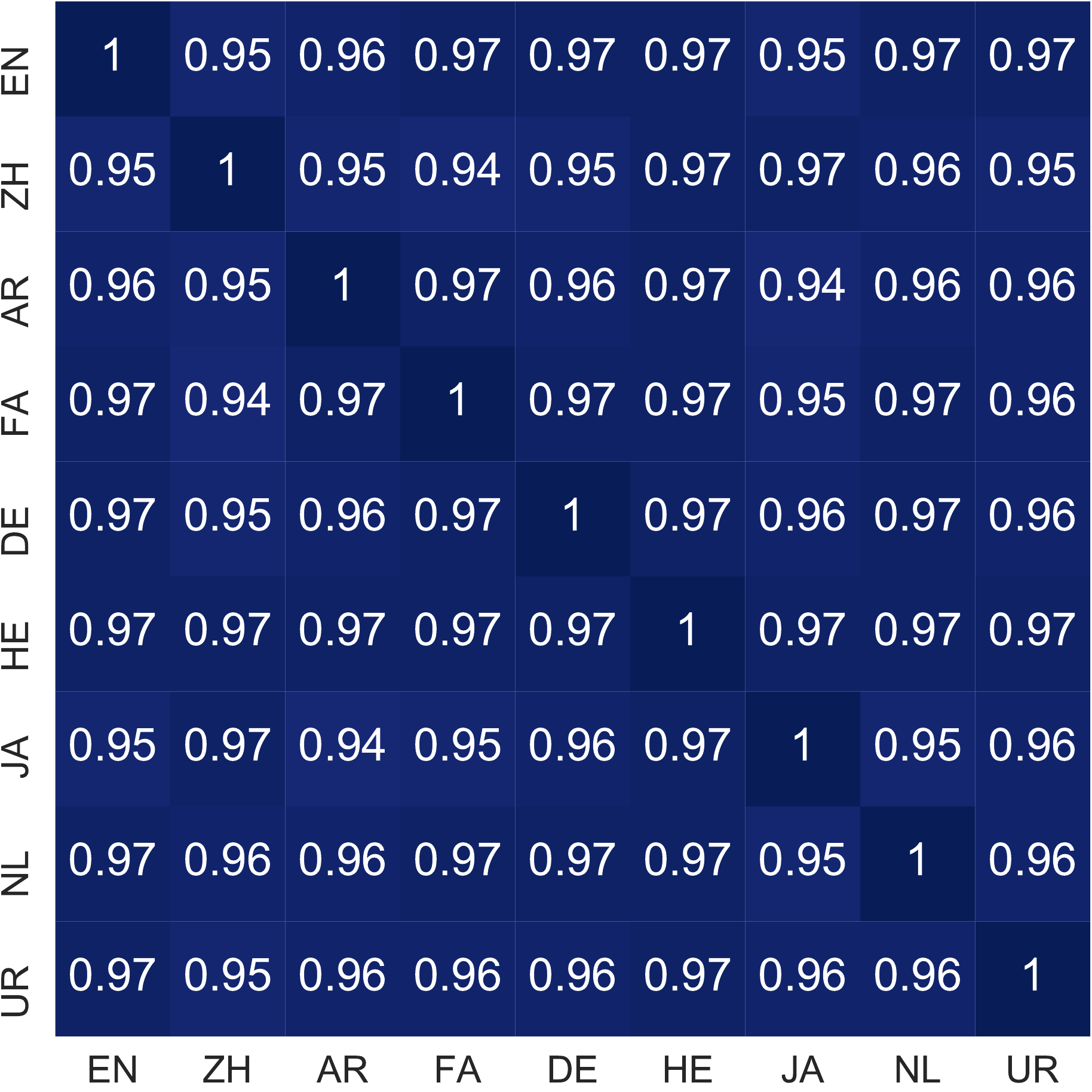}
    \caption{POS-XLM-R}
\end{subfigure}
   \hfill
  \begin{subfigure}{.49\linewidth}
    \centering
    \includegraphics[height=1\linewidth]{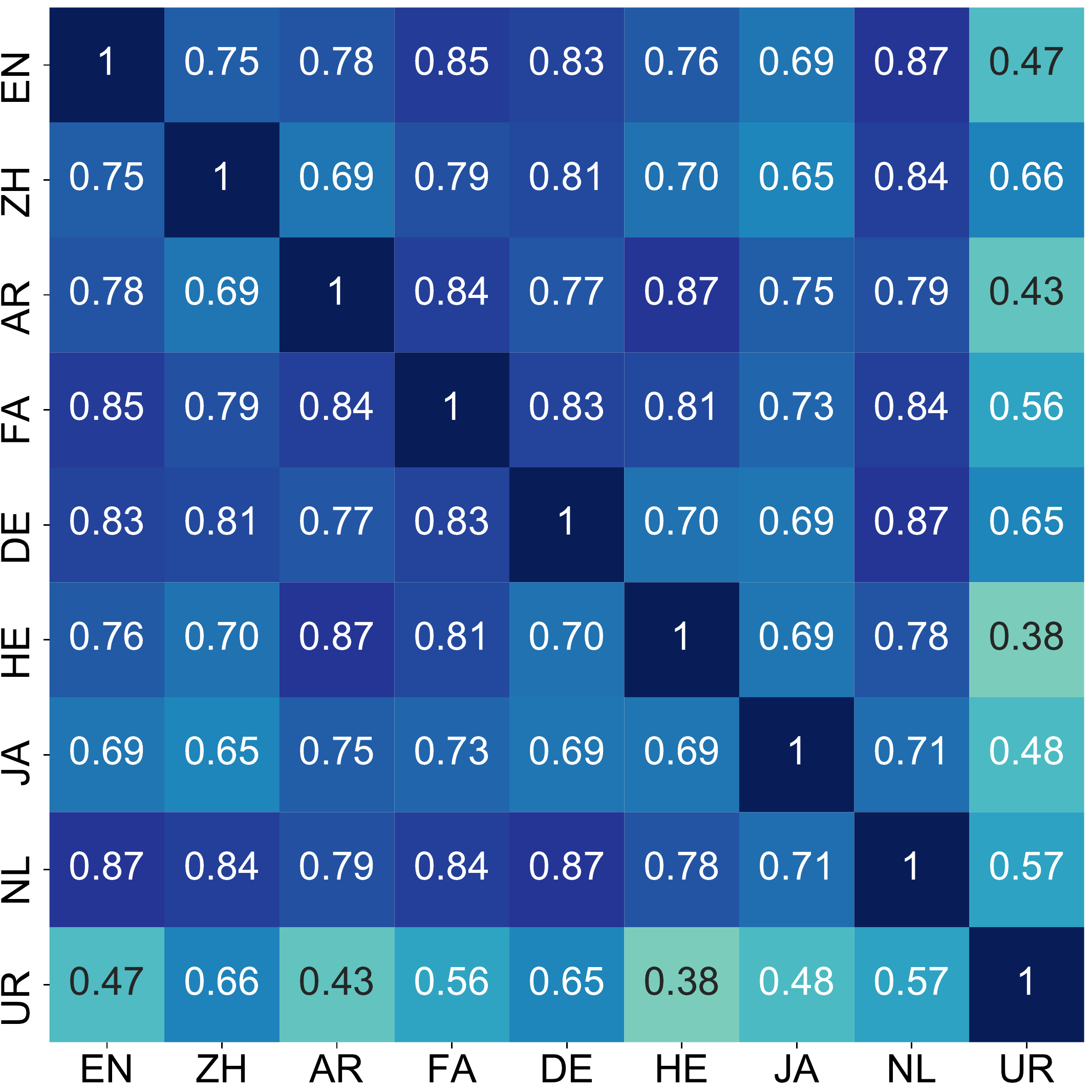}
    \caption{NER-mBERT}
\end{subfigure}
    \hfill
\begin{subfigure}{.49\linewidth}
    \centering
    \includegraphics[height=1\linewidth]{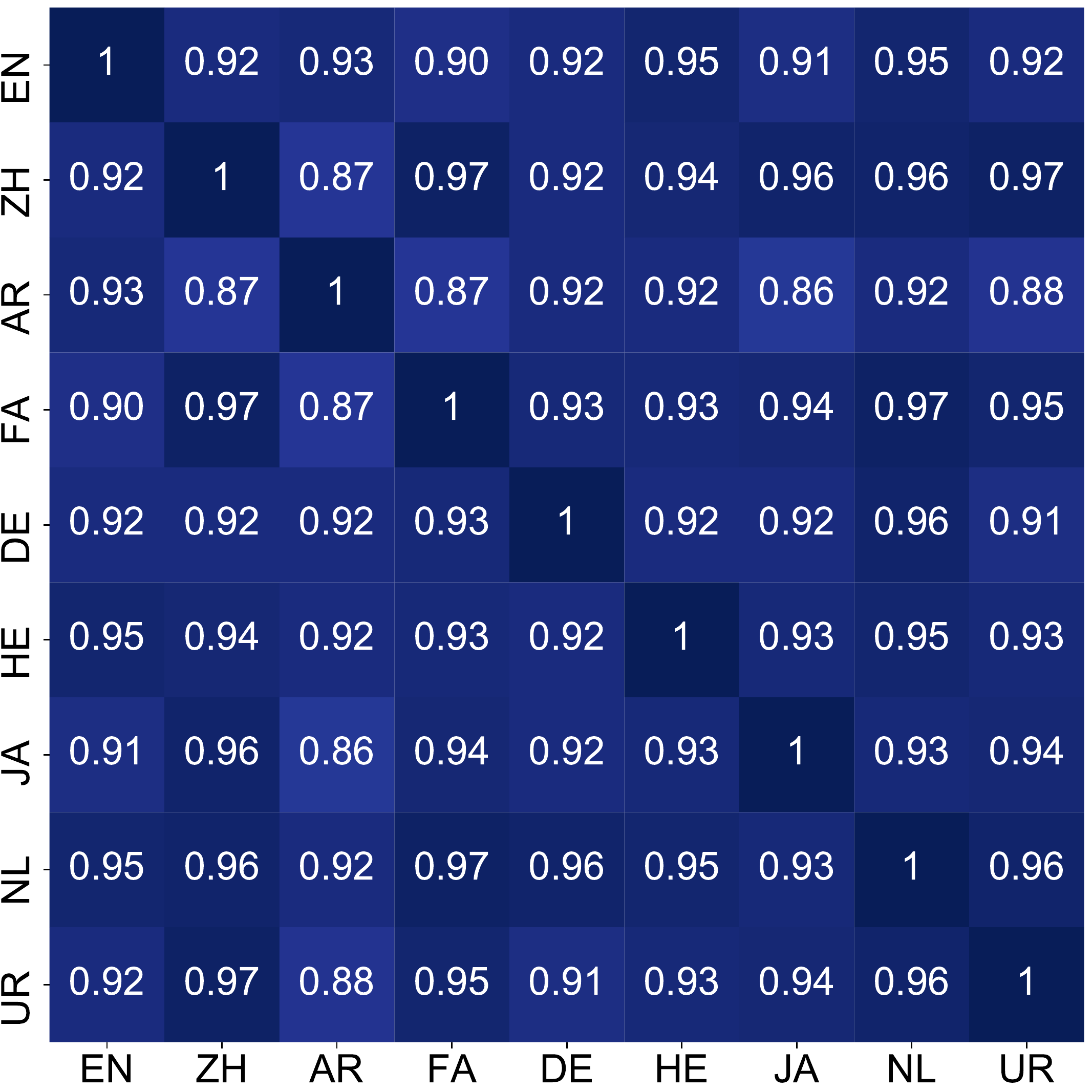}
    \caption{NER-XLM-R}
\end{subfigure}
   \hfill
   
\begin{subfigure}{.49\linewidth}
    \centering
    \includegraphics[height=1\linewidth]{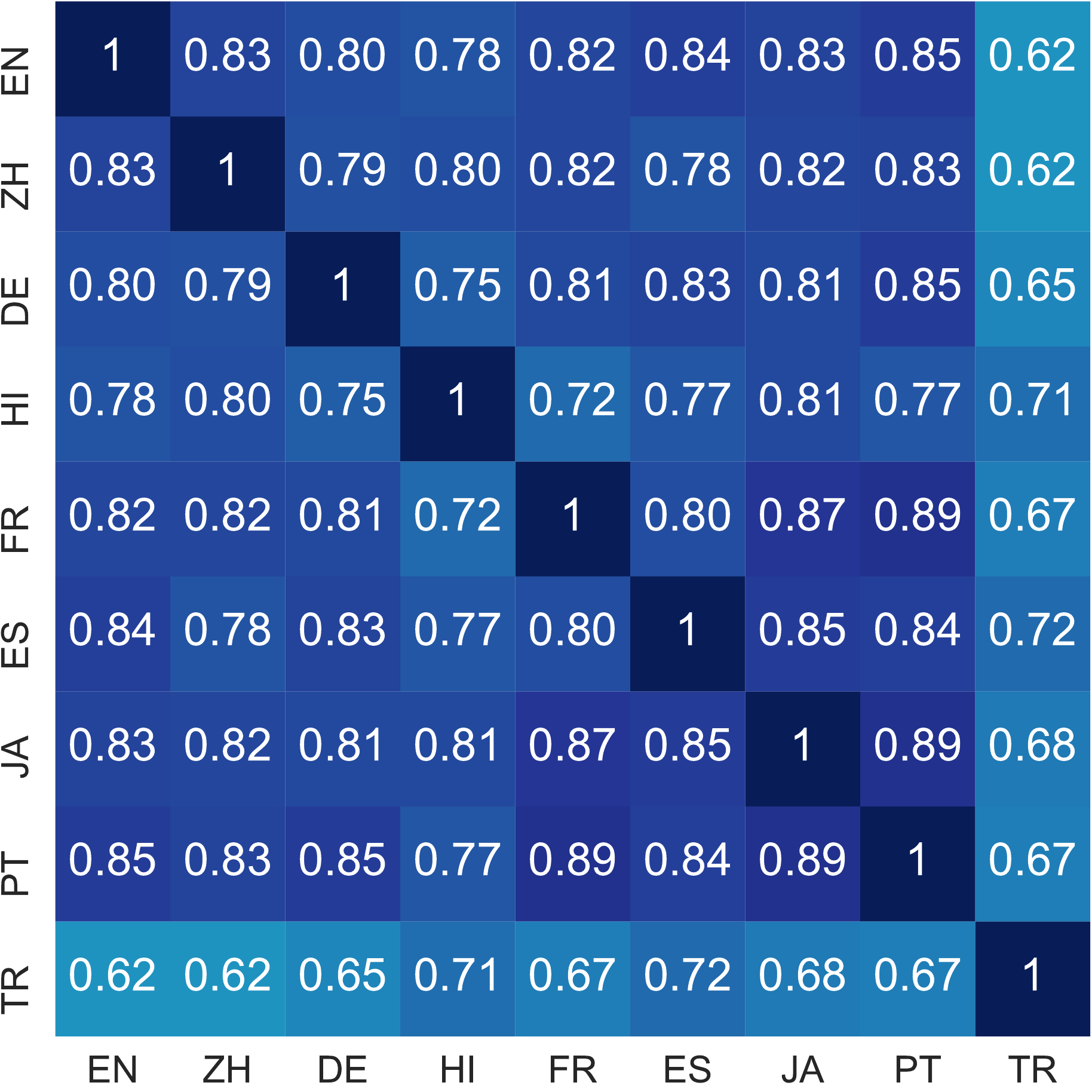}
    \caption{SF-mBERT}
\end{subfigure}
    \hfill
\begin{subfigure}{.49\linewidth}
    \centering
    \includegraphics[height=1\linewidth]{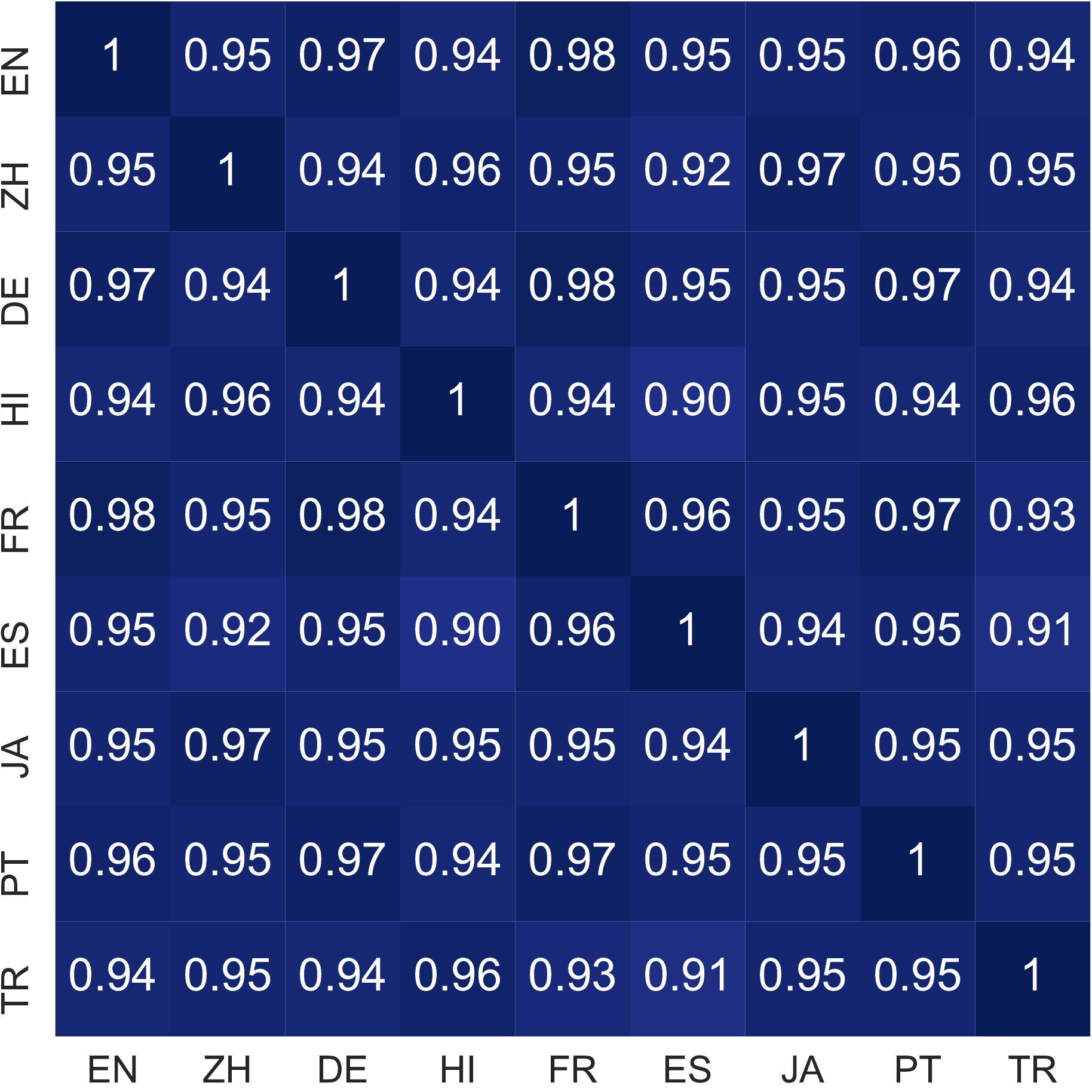}
    \caption{SF-XLM-R}
\end{subfigure}
   \hfill
\caption{Spearman's $\rho$ of head ranking matrices between languages in the POS, NER, and SF tasks. Darker colors indicate higher correlations.}
\label{fig:similarity-head-rankings}
\end{figure}

We show Spearman's rank correlation coefficients (Spearman's $\rho$) between head rankings of each language pair generated by our method on POS, NER, and SF in Figure \ref{fig:similarity-head-rankings}.
The highest-ranked heads largely overlap in all three tasks, while the rankings of unimportant heads vary more in mBERT than XLM-R.

After ranking the attention heads, we fine-tune the model, with the lowest-ranked head in the source language pruned.
We keep increasing the number of heads to prune until it reaches a pre-set limit or when the performance starts to drop.
We limit the number of trials to 12 since the models mostly show improved performance within 12 attempts \footnote{On average 7.52 and 6.58 heads are pruned for POS, 7.54 and 7.28 heads for NER, and 6.19 and 6.31 heads for SF, respectively in mBERT and XLM-R models.}.

\begin{table*}[t]
\centering
\begin{tabular}{|c|l|r|r|r|r|r|r|r|r|}
\hline
\multirow{3}{*}{SL} & \multicolumn{1}{c|}{\multirow{3}{*}{TL}} & \multicolumn{4}{c|}{mBERT}                                                                                              & \multicolumn{4}{c|}{XLM-R}                                                                                              \\ \cline{3-10} 
                    & \multicolumn{1}{c|}{}                    & \multicolumn{2}{c|}{Unpruned}                              & \multicolumn{2}{c|}{Pruned}                                & \multicolumn{2}{c|}{Unpruned}                              & \multicolumn{2}{c|}{Pruned}                                \\ \cline{3-10} 
                    & \multicolumn{1}{c|}{}                    & \multicolumn{1}{c|}{CrLing} & \multicolumn{1}{c|}{MulLing} & \multicolumn{1}{c|}{CrLing} & \multicolumn{1}{c|}{MulLing} & \multicolumn{1}{c|}{CrLing} & \multicolumn{1}{c|}{MulLing} & \multicolumn{1}{c|}{CrLing} & \multicolumn{1}{c|}{MulLing} \\ \hline
\multirow{8}{*}{EN} & ZH                                       & 59.88                       & 95.10                        & \textbf{59.99}              & \textbf{95.31}               & 41.10                       & 95.87                        & \textbf{46.18}              & \textbf{95.99}               \\ \cline{2-10} 
                    & AR                                       & 55.98                       & 95.64                        & \textbf{56.71}              & \textbf{95.68}               & 66.75                       & 96.07                        & \textbf{67.02}              & \textbf{96.13}               \\ \cline{2-10} 
                    & FA                                       & 57.94                       & 94.48                        & \textbf{58.34}              & \textbf{94.81}               & \textbf{66.60}              & 96.85                        & 66.50                       & \textbf{97.09}               \\ \cline{2-10} 
                    & DE                                       & 88.86                       & 94.81                        & \textbf{89.13}              & \textbf{94.94}               & 89.41                       & 94.81                        & \textbf{89.78}              & \textbf{95.19}               \\ \cline{2-10} 
                    & HE                                       & 77.91                       & 96.45                        & \textbf{78.01}              & \textbf{96.58}               & 77.48                       & 97.26                        & \textbf{80.37}              & \textbf{97.30}               \\ \cline{2-10} 
                    & JA                                       & 44.73                       & 96.84                        & \textbf{45.95}              & \textbf{96.97}               & 30.98                       & 97.52                        & \textbf{33.64}              & \textbf{97.62}               \\ \cline{2-10} 
                    & NL                                       & 87.45                       & 96.47                        & \textbf{87.48}              & \textbf{96.69}               & \textbf{88.06}              & \textbf{97.04}               & 88.03                       & 97.02                        \\ \cline{2-10} 
                    & UR                                       & 53.21                       & 91.92                        & \textbf{54.78}              & \textbf{92.17}               & 55.45                       & 92.94                        & \textbf{56.04}              & \textbf{93.07}               \\ \hline
\multirow{8}{*}{ZH} & EN                                       & 55.63                       & 96.52                        & \textbf{57.05}              & \textbf{96.64}               & 42.35                       & 97.19                        & \textbf{43.38}              & \textbf{97.32}               \\ \cline{2-10} 
                    & AR                                       & 38.41                       & 95.62                        & \textbf{41.03}              & \textbf{95.66}               & 36.71                       & 95.99                        & \textbf{38.19}              & \textbf{96.07}               \\ \cline{2-10} 
                    & FA                                       & 43.68                       & 94.55                        & \textbf{45.29}              & \textbf{94.63}               & 33.43                       & 97.07                        & \textbf{34.64}              & \textbf{97.09}               \\ \cline{2-10} 
                    & DE                                       & 63.50                       & 94.62                        & \textbf{64.36}              & \textbf{94.75}               & 46.58                       & 95.06                        & \textbf{47.47}              & \textbf{95.22}               \\ \cline{2-10} 
                    & HE                                       & 57.14                       & 96.51                        & \textbf{57.94}              & \textbf{96.58}               & \textbf{51.26}              & 97.06                        & 50.42                       & \textbf{97.19}               \\ \cline{2-10} 
                    & JA                                       & 43.63                       & 96.73                        & \textbf{44.69}              & \textbf{97.01}               & 49.12                       & 97.32                        & \textbf{49.74}              & \textbf{97.34}               \\ \cline{2-10} 
                    & NL                                       & 59.95                       & 96.78                        & \textbf{61.10}              & \textbf{96.97}               & 40.78                       & 97.30                        & \textbf{42.50}              & \textbf{97.43}               \\ \cline{2-10} 
                    & UR                                       & 43.82                       & 92.21                        & \textbf{44.07}              & \textbf{92.26}               & \textbf{30.08}              & 92.90                        & 29.26                       & \textbf{93.01}               \\ \hline
\multirow{8}{*}{AR} & EN                                       & 54.77                       & 96.50                        & \textbf{56.90}              & \textbf{96.53}               & 61.73                       & 97.21                        & \textbf{63.63}              & \textbf{97.31}               \\ \cline{2-10} 
                    & ZH                                       & 46.19                       & 95.16                        & \textbf{47.14}              & \textbf{95.31}               & 25.12                       & 95.16                        & \textbf{34.71}              & \textbf{96.04}               \\ \cline{2-10} 
                    & FA                                       & 63.82                       & 94.52                        & \textbf{64.02}              & \textbf{94.64}               & 70.92                       & 97.15                        & \textbf{71.55}              & \textbf{97.20}               \\ \cline{2-10} 
                    & DE                                       & 56.88                       & 94.82                        & \textbf{57.85}              & \textbf{94.98}               & 65.21                       & 95.16                        & \textbf{68.28}              & \textbf{95.29}               \\ \cline{2-10} 
                    & HE                                       & 60.33                       & 96.44                        & \textbf{61.88}              & \textbf{96.70}               & 67.45                       & 97.23                        & \textbf{67.72}              & \textbf{97.34}               \\ \cline{2-10} 
                    & JA                                       & \textbf{44.32}              & 97.02                        & 44.18                       & \textbf{97.15}               & 22.11                       & 97.52                        & \textbf{29.21}              & \textbf{97.65}               \\ \cline{2-10} 
                    & NL                                       & 58.86                       & 96.87                        & \textbf{60.31}              & \textbf{97.03}               & 62.93                       & 96.87                        & \textbf{64.80}              & \textbf{97.50}               \\ \cline{2-10} 
                    & UR                                       & 49.31                       & 92.00                        & \textbf{49.76}              & \textbf{92.16}               & 54.79                       & 92.74                        & \textbf{56.06}              & \textbf{92.88}               \\ \cline{2-10} \hline
\end{tabular}
\caption{F-1 scores of mBERT and XLM on POS. SL and TL refer to source and target languages and CrLing and MulLing stand for cross-lingual and multi-lingual settings, respectively. Unpruned results are produced by the full models and pruned results are the best scores each model produces with up to 12 lowest-ranked heads pruned. The higher performance in each pair of pruned and unpruned experiments is in bold.}
\label{tbl:results-pos-single-source}
\end{table*}

\begin{table*}[t]
\centering
\begin{tabular}{|c|c|r|r|r|r|r|r|r|r|}
\hline
\multirow{3}{*}{SL} & \multirow{3}{*}{TL} & \multicolumn{4}{c|}{mBERT}                                                                                              & \multicolumn{4}{c|}{XLM-R}                                                                                              \\ \cline{3-10} 
                    &                     & \multicolumn{2}{c|}{Unpruned}                              & \multicolumn{2}{c|}{Pruned}                                & \multicolumn{2}{c|}{Unpruned}                              & \multicolumn{2}{c|}{Pruned}                                \\ \cline{3-10} 
                    &                     & \multicolumn{1}{c|}{CrLing} & \multicolumn{1}{c|}{MulLing} & \multicolumn{1}{c|}{CrLing} & \multicolumn{1}{c|}{MulLing} & \multicolumn{1}{c|}{CrLing} & \multicolumn{1}{c|}{MulLing} & \multicolumn{1}{c|}{CrLing} & \multicolumn{1}{c|}{MulLing} \\ \hline
\multirow{8}{*}{EN} & ZH                  & 47.64                       & 93.24                        & \textbf{51.61}              & \textbf{93.71}               & 29.97                       & 90.99                        & \textbf{32.33}              & \textbf{91.11}               \\ \cline{2-10} 
                    & AR                  & 38.81                       & 70.55                        & \textbf{38.93}              & \textbf{73.32}               & 41.21                       & 71.77                        & \textbf{43.78}              & \textbf{74.28}               \\ \cline{2-10} 
                    & FA                  & \textbf{40.12}              & 96.70                        & 39.81                       & \textbf{96.97}               & 54.90                       & 96.62                        & \textbf{55.72}              & \textbf{96.98}               \\ \cline{2-10} 
                    & DE                  & 56.43                       & 79.11                        & \textbf{58.27}              & \textbf{79.19}               & 63.71                       & 82.31                        & \textbf{66.48}              & \textbf{83.10}               \\ \cline{2-10} 
                    & HE                  & \textbf{46.92}              & \textbf{89.18}               & 46.55                       & 88.49                        & \textbf{56.96}              & 88.02                        & 56.87                       & \textbf{89.67}               \\ \cline{2-10} 
                    & JA                  & 42.45                       & \textbf{84.91}               & \textbf{44.14}              & 84.34                        & 33.87                       & 81.48                        & \textbf{37.88}              & \textbf{82.35}               \\ \cline{2-10} 
                    & NL                  & 64.51                       & 84.90                        & \textbf{65.56}              & \textbf{85.17}               & 77.15                       & 90.21                        & \textbf{77.66}              & \textbf{90.38}               \\ \cline{2-10} 
                    & UR                  & 37.34                       & \textbf{99.29}               & \textbf{40.60}              & 99.22                        & 58.25                       & \textbf{99.15}               & \textbf{58.68}              & 99.07                        \\ \hline
\multirow{8}{*}{ZH} & EN                  & 38.58                       & 87.65                        & \textbf{41.40}              & \textbf{87.99}               & 56.40                       & 90.72                        & \textbf{58.55}              & \textbf{91.05}               \\ \cline{2-10} 
                    & AR                  & 36.43                       & 72.27                        & \textbf{36.99}              & \textbf{72.86}               & 34.31                       & 74.84                        & \textbf{36.11}              & \textbf{75.68}               \\ \cline{2-10} 
                    & FA                  & 45.68                       & 96.21                        & \textbf{46.57}              & \textbf{96.23}               & \textbf{51.60}              & 95.63                        & 51.51                       & \textbf{95.66}               \\ \cline{2-10} 
                    & DE                  & 29.07                       & \textbf{79.04}               & \textbf{33.81}              & 78.67                        & \textbf{56.22}              & 82.33                        & 55.51                       & \textbf{82.54}               \\ \cline{2-10} 
                    & HE                  & 47.14                       & 88.20                        & \textbf{47.68}              & \textbf{89.35}               & 48.52                       & 85.95                        & \textbf{48.94}              & \textbf{87.79}               \\ \cline{2-10} 
                    & JA                  & 49.21                       & 82.02                        & \textbf{51.69}              & \textbf{83.20}               & 46.18                       & 80.19                        & \textbf{47.06}              & \textbf{82.63}               \\ \cline{2-10} 
                    & NL                  & 29.75                       & 84.61                        & \textbf{31.46}              & \textbf{85.28}               & 49.59                       & 89.56                        & \textbf{52.27}              & \textbf{90.56}               \\ \cline{2-10} 
                    & UR                  & 44.61                       & 99.26                        & \textbf{46.33}              & \textbf{99.28}               & 48.98                       & 98.99                        & \textbf{55.95}              & \textbf{99.10}               \\ \hline
\multirow{8}{*}{AR} & EN                  & 19.29                       & \textbf{87.86}               & \textbf{20.07}              & 87.82                        & \textbf{51.33}              & 90.37                        & 51.00                       & \textbf{91.01}               \\ \cline{2-10} 
                    & ZH                  & \textbf{41.70}              & 93.46                        & 40.43                       & \textbf{93.54}               & 25.78                       & 90.51                        & \textbf{31.03}              & \textbf{91.00}               \\ \cline{2-10} 
                    & FA                  & 46.57                       & 96.82                        & \textbf{46.87}              & \textbf{96.87}               & \textbf{53.35}              & 96.55                        & 52.60                       & \textbf{96.74}               \\ \cline{2-10} 
                    & DE                  & 24.47                       & 75.78                        & \textbf{25.62}              & \textbf{78.04}               & \textbf{50.87}              & 82.63                        & 50.00                       & \textbf{82.73}               \\ \cline{2-10} 
                    & HE                  & \textbf{47.15}              & 86.77                        & 46.72                       & \textbf{87.64}               & 49.52                       & 87.37                        & \textbf{50.85}              & \textbf{89.28}               \\ \cline{2-10} 
                    & JA                  & 41.49                       & 79.90                        & \textbf{42.11}              & \textbf{83.17}               & 36.98                       & \textbf{81.72}               & \textbf{38.87}              & 80.92                        \\ \cline{2-10} 
                    & NL                  & 26.00                       & 84.83                        & \textbf{26.34}              & \textbf{85.24}               & \textbf{49.27}              & 90.73                        & 48.87                       & \textbf{91.11}               \\ \cline{2-10} 
                    & UR                  & \textbf{46.47}              & 99.26                        & 45.66                       & \textbf{99.31}               & 48.48                       & 99.10                        & \textbf{53.51}              & \textbf{99.15}               \\ \hline
\end{tabular}
\caption{F-1 scores of mBERT and XLM on NER. SL and TL refer to source and target languages and CrLing and MulLing stand for cross-lingual and multi-lingual settings, respectively. Unpruned results are produced by the full models and pruned results are the best scores each model produces with up to 12 lowest-ranked heads pruned.}
\label{tbl:results-ner-single-source}
\end{table*}

\begin{table*}[]
\centering
\begin{tabular}{|c|c|r|r|r|r|r|r|r|r|}
\hline
\multirow{3}{*}{SL} & \multirow{3}{*}{TL}     & \multicolumn{4}{c|}{mBERT}                                                                                              & \multicolumn{4}{c|}{XLM-R}                                                                                              \\ \cline{3-10} 
                    &                         & \multicolumn{2}{c|}{Unpruned}                              & \multicolumn{2}{c|}{Pruned}                                & \multicolumn{2}{c|}{Unpruned}                              & \multicolumn{2}{c|}{Pruned}                                \\ \cline{3-10} 
                    &                         & \multicolumn{1}{c|}{CrLing} & \multicolumn{1}{c|}{MulLing} & \multicolumn{1}{c|}{CrLing} & \multicolumn{1}{c|}{MulLing} & \multicolumn{1}{c|}{CrLing} & \multicolumn{1}{c|}{MulLing} & \multicolumn{1}{c|}{CrLing} & \multicolumn{1}{c|}{MulLing} \\ \hline
\multirow{9}{*}{EN} & ZH                      & 69.83                       & 94.11                        & \textbf{71.84}              & \textbf{94.25}               & 62.58                       & 93.97                        & \textbf{67.98}              & \textbf{94.29}               \\ \cline{2-10} 
                    & DE                      & 60.69                       & 94.60                        & \textbf{66.97}              & \textbf{94.95}               & 82.85                       & 94.81                        & \textbf{83.50}              & \textbf{95.35}               \\ \cline{2-10} 
                    & HI                      & 44.28                       & 85.93                        & \textbf{45.84}              & \textbf{87.08}               & 58.32                       & 86.72                        & \textbf{66.39}              & \textbf{87.16}               \\ \cline{2-10} 
                    & FR                      & 60.44                       & 93.96                        & \textbf{67.13}              & \textbf{94.18}               & 76.53                       & 93.51                        & \textbf{77.59}              & \textbf{93.77}               \\ \cline{2-10} 
                    & ES                      & 72.27                       & 87.71                        & \textbf{73.96}              & \textbf{88.17}               & 81.70                       & \textbf{89.10}               & \textbf{81.88}              & 88.83                        \\ \cline{2-10} 
                    & JA                      & 68.28                       & 93.73                        & \textbf{68.32}              & \textbf{93.78}               & 32.39                       & 93.65                        & \textbf{36.68}              & \textbf{93.71}               \\ \cline{2-10} 
                    & PT                      & 59.37                       & \textbf{90.83}               & \textbf{63.23}              & 90.82                        & 77.42                       & 90.76                        & \textbf{77.54}              & \textbf{91.24}               \\ \cline{2-10} 
                    & TR                      & 28.11                       & 83.41                        & \textbf{32.21}              & \textbf{84.31}               & 45.91                       & 83.20                        & \textbf{52.64}              & \textbf{84.30}               \\ \cline{2-10} 
                    & \multicolumn{1}{l|}{EN} & \multicolumn{2}{c|}{\textbf{95.43}}                                      & \multicolumn{2}{c|}{95.27}                                      & \multicolumn{2}{c|}{94.59}                                      & \multicolumn{2}{c|}{\textbf{94.87}}                                      \\ \hline
\end{tabular}
\caption{Slot F-1 scores on the MultiAtis++ corpus. CrLing and MulLing refer to cross-lingual and multi-lingual settings, respectively. SL and TL refer to source and target languages, respectively. English mono-lingual results are reported for validity check purposes.}
\label{tbl:results-slot-filling}
\end{table*}

\section{Experiments and Analysis}

This section displays and explains experimental results on cross-lingual and multi-lingual POS, NER, and SF tasks. Training sets in target languages are not used to train the model under the cross-lingual setting.
Our experiments are based on the Huggingface \cite{huggingface} implementations of mBERT and XLM-R.
Specifically, we use the pre-trained bert-base-multilingual-cased and xlm-roberta-base models for their comparable model sizes.
The models are fine-tuned for 3 epochs with a learning rate of 5e-5 in all the experiments.
We use the official dataset splits and load training instances with sequential data samplers, so the reported evaluation scores are robust to randomness.

\subsection{POS}
Table \ref{tbl:results-pos-single-source} shows the evaluation scores on POS with three source language choices.
In the majority (88 out of 96 pairs) of experiments, pruning up to 12 attention heads improves mBERT and XLM-R performance.
Results are comparable in the other 8 experiments with and without head pruning.
Average F-1 score improvements are 0.91 for mBERT and 1.78 for XLM-R in cross-lingual tasks, and 0.15 for mBERT and 0.17 for XLM-R in multi-lingual tasks.
These results support that pruning heads generally has positive effects on model performance in cross-lingual and multi-lingual tasks, and that our method correctly ranks the heads.

Consistent with \citet{xlm-r-orig}, XLM-R usually outperforms mBERT, with exceptions in cross-lingual experiments where ZH and JA datasets are involved.
Word segmentation in ZH and JA is different from the other languages we choose, e.g. words are not separated by white spaces and unpaired adjacent word pieces often make up a new word.
As XLM-R applies the SentencePiece tokenization method \cite{sentence-piece}, it is more likely to detect wrong word boundaries and make improper predictions than mBERT in cross-lingual experiments involving ZH or JA datasets.
We note that the performance improvements are solid regardless of the source language selection and severe differences of training data sizes in EN, ZH, and AR.
This demonstrates the correctness of the head rankings our method generates and that the important attention heads for a task are almost language invariant.

We also examine to what extent the score improvements are affected by the relationships between source and target languages, e.g. language families, URIEL language distance scores \cite{uriel}, and the similarity of the head ranking matrices.
There are three non-exclusive clusters of language families (containing more than one language) in our choice of languages, namely Indo-European (IE), Germanic, and Semitic languages.
Average score improvements between models with and without head pruning are 0.40 (IE), 0.16 (Germanic), and 0.91 (Semitic) for mBERT and 0.19 (IE), 0.18 (Germanic), and 0.19 (Semitic) for XLM-R.
In comparison, the overall average score improvements are 0.53 for mBERT and 0.97 for XLM-R.
Despite the generally higher performance of models when the source and target languages are in the same family,
the score improvements by pruning heads are not necessarily associated with language families.
Additionally, we use Spearman's $\rho$ to measure the correlations between improved F-1 scores and URIEL language distances.
The correlation scores are 0.11 (cross-lingual) and 0.12 (multi-lingual) for mBERT, and -0.40 (cross-lingual) and 0.23 (multi-lingual) for XLM-R. 
Similarly, the Spearman's $\rho$ between score improvements and similarities in head ranking matrices shown in Figure \ref{fig:similarity-head-rankings} are -0.34 (cross-lingual) and 0.25 (multi-lingual) for mBERT, and -0.52 (cross-lingual) and -0.10 (multi-lingual) for XLM-R.
This indicate that except in the cross-lingual XLM-R model which faces word segmentation issues on ZH or JA experiments, pruning attention heads improves model performance regardless of the distances between source and target languages.
Thus our findings are potentially applicable to all cross-lingual and multi-lingual POS tasks.

\subsection{NER}
As Table \ref{tbl:results-ner-single-source} shows, pruning attention heads generally has positive effects on our cross-lingual and multi-lingual NER models.
Even in the multi-lingual AR-UR experiment where the full mBERT model achieves an F-1 score of 99.26,
the score is raised to 99.31 by pruning heads.
Scores are comparable with and without head pruning in the 19 cases where model performances are not improved.
This also lends support to the specialized role of important attention heads and the consistency of head rankings across languages.
In NER experiments, performance drops mostly happen when the source and target languages are from different families.
This is likely caused by the difference between named entity (NE) representations across language families.
We show in Section \ref{sct:multi-source-case-study} that the gap is largely bridged when a language from the same family as the target language is added to the source languages. 

Average score improvements are comparable on mBERT (0.81 under cross-lingual and 0.31 under multi-lingual settings) and XLM-R (1.08 under cross-lingual and 0.67 under multi-lingual settings) in NER experiments.
The results indicate that the performance improvements introduced by head-pruning are not sensitive to the pre-training corpora of models.
The correlations between F-1 score improvements and URIEL language distances are small, with Spearman's $\rho$ of -0.05 (cross-lingual) and -0.27 (multi-lingual) for mBERT and 0.10 (cross-lingual) and 0.12 (multi-lingual) for XLM-R.
Similarities between head ranking matrices do not greatly affect score improvements either, the Spearman's $\rho$ of which are -0.08 (cross-lingual) and 0.06 (multi-lingual) for mBERT and 0.05 (cross-lingual) and 0.12 (multi-lingual) for XLM-R.
The findings in POS and NER experiments are consistent, supporting our hypothesis that important heads for a task are shared by arbitrary source-target language selections.

\subsection{Slot Filling}
We report SF evaluation results in Table \ref{tbl:results-slot-filling}.
In 31 out of 34 pairs of experiments, pruning up to 12 heads results in performance improvements, while the scores are comparable in the other three cases.
These results agree with those in POS and NER experiments, showing that only a subset of heads in each model makes key contributions to cross-lingual or multi-lingual tasks.

We also evaluate the correlations between score changes and the closeness of source and target languages.
In terms of URIEL language distances, the Spearman's $\rho$ are 0.69 (cross-lingual) and 0.14 (multi-lingual) for mBERT and -0.59 (cross-lingual) and 0.14 (multi-lingual) for XLM-R.
The coefficients are -0.25 (cross-lingual) and -0.73 (multi-lingual) for mBERT and -0.70 (cross-lingual) and -0.14 (multi-lingual) between score improvements and similarities in head ranking matrices.
While these coefficients are generally higher than those in POS and NER evaluations, their p-values are also high (0.55 to 0.74), indicating the correlations between the score changes and source-target language closeness are not statistically significant. \footnote{The p-values for all the other Spearman's $\rho$ we report are lower than 0.01, showing that those correlation scores are statistically significant.}

\section{Discussions} \label{sct:case-study}
In this section, we perform case studies to confirm the validity of our head ranking method.
We also illustrate the extensibility of the knowledge we learn from the main experiments to a wider range of settings, e.g. when the training dataset is limited in size or constructed over multiple source languages.

\subsection{Correctness of Head Rankings}
\begin{table}[]
\centering
\begin{tabular}{|l|r|r|r|r|}
\hline
\multicolumn{5}{|c|}{NER}\\\hline
\multicolumn{1}{|c|}{}                     & \multicolumn{2}{c|}{Max-Pruning}                                                                               & \multicolumn{2}{c|}{Rand-Pruning}    
\\ \cline{2-5} 
\multicolumn{1}{|c|}{\multirow{-2}{*}{TL}} & \multicolumn{1}{c|}{CrLing}                          & \multicolumn{1}{c|}{MulLing}                         & \multicolumn{1}{c|}{CrLing}                          & \multicolumn{1}{c|}{MulLing}                         \\ \hline
ZH                                         & \cellcolor[HTML]{6699ff}{\color[HTML]{FFFFFF} -1.74} & \cellcolor[HTML]{ff6666}{\color[HTML]{FFFFFF} +0.08} & \cellcolor[HTML]{6699ff}{\color[HTML]{FFFFFF} -2.44} & \cellcolor[HTML]{ff6666}{\color[HTML]{FFFFFF} +0.26} \\ \hline
AR                                         & \cellcolor[HTML]{6699ff}{\color[HTML]{FFFFFF} -3.17} & \cellcolor[HTML]{6699ff}{\color[HTML]{FFFFFF} -2.42} & \cellcolor[HTML]{6699ff}{\color[HTML]{FFFFFF} -2.09} & \cellcolor[HTML]{6699ff}{\color[HTML]{FFFFFF} -0.43} \\ \hline
DE                                         & \cellcolor[HTML]{ff6666}{\color[HTML]{FFFFFF} +0.88} & \cellcolor[HTML]{6699ff}{\color[HTML]{FFFFFF} -0.62} & \cellcolor[HTML]{ff6666}{\color[HTML]{FFFFFF} +0.57} & \cellcolor[HTML]{6699ff}{\color[HTML]{FFFFFF} -0.38} \\ \hline
NL                                         & \cellcolor[HTML]{6699ff}{\color[HTML]{FFFFFF} -2.76} & \cellcolor[HTML]{6699ff}{\color[HTML]{FFFFFF} -0.23} & \cellcolor[HTML]{ff6666}{\color[HTML]{FFFFFF} +0.29} & \cellcolor[HTML]{ff6666}{\color[HTML]{FFFFFF} +0.36} \\ \hline
FA                                         & \cellcolor[HTML]{6699ff}{\color[HTML]{FFFFFF} -0.86} & \cellcolor[HTML]{6699ff}{\color[HTML]{FFFFFF} -0.31} & \cellcolor[HTML]{6699ff}{\color[HTML]{FFFFFF} -2.52} & \cellcolor[HTML]{6699ff}{\color[HTML]{FFFFFF} -0.74} \\ \hline
HE                                         & \cellcolor[HTML]{6699ff}{\color[HTML]{FFFFFF} -2.50} & \cellcolor[HTML]{6699ff}{\color[HTML]{FFFFFF} -2.15} & \cellcolor[HTML]{6699ff}{\color[HTML]{FFFFFF} -0.49} & \cellcolor[HTML]{6699ff}{\color[HTML]{FFFFFF} -4.21} \\ \hline
JA                                         & \cellcolor[HTML]{6699ff}{\color[HTML]{FFFFFF} -1.48} & \cellcolor[HTML]{6699ff}{\color[HTML]{FFFFFF} -1.08} & \cellcolor[HTML]{6699ff}{\color[HTML]{FFFFFF} -2.65} & \cellcolor[HTML]{6699ff}{\color[HTML]{FFFFFF} -2.40} \\ \hline
UR                                         & \cellcolor[HTML]{6699ff}{\color[HTML]{FFFFFF} -0.15} & \cellcolor[HTML]{6699ff}{\color[HTML]{FFFFFF} -0.10} & \cellcolor[HTML]{6699ff}{\color[HTML]{FFFFFF} -0.60} & \cellcolor[HTML]{6699ff}{\color[HTML]{FFFFFF} -0.12} \\ \hline
\multicolumn{5}{|c|}{POS}\\
\hline
\multicolumn{1}{|c|}{}                     & \multicolumn{2}{c|}{Max-Mask}                                                                               & \multicolumn{2}{c|}{Rand-Mask}                                                                              \\ \cline{2-5} 
\multicolumn{1}{|c|}{\multirow{-2}{*}{TL}} & \multicolumn{1}{c|}{CrLing}                          & \multicolumn{1}{c|}{MulLing}                         & \multicolumn{1}{c|}{CrLing}                          & \multicolumn{1}{c|}{MulLing}                         \\ \hline
ZH                                         & \cellcolor[HTML]{ff6666}{\color[HTML]{FFFFFF} +0.03} & \cellcolor[HTML]{6699ff}{\color[HTML]{FFFFFF} -0.39} & \cellcolor[HTML]{6699ff}{\color[HTML]{FFFFFF} -0.14} & \cellcolor[HTML]{6699ff}{\color[HTML]{FFFFFF} -0.20} \\ \hline
AR                                         & \cellcolor[HTML]{6699ff}{\color[HTML]{FFFFFF} -0.65} & \cellcolor[HTML]{6699ff}{\color[HTML]{FFFFFF} -0.04} & \cellcolor[HTML]{6699ff}{\color[HTML]{FFFFFF} -0.66} & \cellcolor[HTML]{6699ff}{\color[HTML]{FFFFFF} -0.12} \\ \hline
DE                                         & \cellcolor[HTML]{6699ff}{\color[HTML]{FFFFFF} -0.64} & \cellcolor[HTML]{6699ff}{\color[HTML]{FFFFFF} -0.04} & \cellcolor[HTML]{6699ff}{\color[HTML]{FFFFFF} -0.64} & \cellcolor[HTML]{6699ff}{\color[HTML]{FFFFFF} -0.14} \\ \hline
NL                                         & \cellcolor[HTML]{6699ff}{\color[HTML]{FFFFFF} -0.13} & \cellcolor[HTML]{6699ff}{\color[HTML]{FFFFFF} -0.13} & \cellcolor[HTML]{6699ff}{\color[HTML]{FFFFFF} -0.11} & \cellcolor[HTML]{6699ff}{\color[HTML]{FFFFFF} -0.16} \\ \hline
FA                                         & \cellcolor[HTML]{6699ff}{\color[HTML]{FFFFFF} -0.75} & \cellcolor[HTML]{6699ff}{\color[HTML]{FFFFFF} -0.03} & \cellcolor[HTML]{6699ff}{\color[HTML]{FFFFFF} -0.53} & \cellcolor[HTML]{6699ff}{\color[HTML]{FFFFFF} -0.25} \\ \hline
HE                                         & \cellcolor[HTML]{6699ff}{\color[HTML]{FFFFFF} -1.27} & \cellcolor[HTML]{6699ff}{\color[HTML]{FFFFFF} -0.28} & \cellcolor[HTML]{6699ff}{\color[HTML]{FFFFFF} -1.06} & \cellcolor[HTML]{ff6666}{\color[HTML]{FFFFFF} +0.05} \\ \hline
JA                                         & \cellcolor[HTML]{6699ff}{\color[HTML]{FFFFFF} -22.29} & \cellcolor[HTML]{6699ff}{\color[HTML]{FFFFFF} -0.05} & \cellcolor[HTML]{6699ff}{\color[HTML]{FFFFFF} -1.23} & \cellcolor[HTML]{6699ff}{\color[HTML]{FFFFFF} -0.05} \\ \hline
UR                                         & \cellcolor[HTML]{6699ff}{\color[HTML]{FFFFFF} -1.78} & \cellcolor[HTML]{6699ff}{\color[HTML]{FFFFFF} -0.11} & \cellcolor[HTML]{6699ff}{\color[HTML]{FFFFFF} -0.77} & \cellcolor[HTML]{6699ff}{\color[HTML]{FFFFFF} -0.07} \\ \hline
\end{tabular}
\caption{F-1 score differences from the full mBERT model on NER (upper) and POS (lower) by pruning highest ranked (Max-Pruning) or random (Rand-Pruning) heads in the ranking matrices. The source language is EN. Blue and red cells indicate score drops and improvements, respectively.}
\label{tbl:validity-discussion}
\end{table}
We evaluate the correctness of our head ranking method through comparisons between results in Tables \ref{tbl:results-pos-single-source} and \ref{tbl:results-ner-single-source} and those produced by pruning (1) randomly sampled heads and (2) highest ranked heads.
Specifically, we repeat the head-pruning experiments with mBERT on NER and POS using EN as the source language and display the score differences from the the full models in Table \ref{tbl:validity-discussion}.
Same as in the main experiments, we pick the best score from pruning 1 to 12 heads in each experiment.
A random seed of 42 is used for sampling attention heads to prune under the random sampling setting.

In 14 out of 16 NER experiments, pruning the heads ranked highest by our method results in noticeable performance drops compared to the full model.
Consistently, pruning the highest-ranked attention heads harms the performance of mBERT in 15 out of 16 POS experiments.
Though score changes are slightly positive for cross-lingual EN-DE and multi-lingual EN-ZH NER tasks and in the cross-lingual EN-ZH POS experiment, improvements introduced by pruning lowest-ranked heads are more significant, as Table \ref{tbl:results-pos-single-source} and Table \ref{tbl:results-ner-single-source} show.
Pruning random attention heads also has mainly negative effects on the performance of mBERT.
These results indicate that while pruning attention heads potentially boosts the performance of models, reasonably choosing the heads to prune is important.
Our gradient-based method properly ranks the heads by their priority to prune.

\subsection{Multiple Source Languages} \label{sct:multi-source-case-study}
Training cross-lingual models on multiple source languages is a practical way to improve their performance, due to enlarged training data size and supervision from source-target languages closer to each other \cite{multi-source-1,multi-source-2,multi-source-3,multi-source-4,multi-source-5}.
We also explore the effects of pruning attention heads under the multi-source settings.
In this section, we experiment with mBERT on EN, DE, AR, HE, and ZH datasets for both NER and POS tasks.
These languages fall into three mutually exclusive language families, enabling our analysis on the influence of training cross-lingual models with source languages belonging to the same family as the target language.
Similar to related research, the model is fine-tuned on the concatenation of training datasets in all the languages but the one on which the model is tested.

\begin{table}[t]
\centering
\begin{tabular}{|c|l|l|l|l|l|}
\hline
\multicolumn{6}{|c|}{NER}\\
\hline
\multicolumn{1}{|l|}{} & \multicolumn{1}{c|}{EN} & \multicolumn{1}{c|}{DE} & \multicolumn{1}{c|}{AR} & \multicolumn{1}{c|}{HE} & \multicolumn{1}{c|}{ZH} \\ \hline
FL                     & 60.77                   & 59.16                   & 35.90                   & 51.19                   & 44.18                   \\ \hline
MD                     & 62.63                   & 61.10                   & 40.78                   & 55.15                   & \textbf{47.59}          \\ \hline
SD                     & 63.38                   & 61.66                   & \textbf{41.53}          & 54.20                   & 47.08                   \\ \hline
EC                     & \textbf{64.63}          & \textbf{61.71}          & 40.78                   & \textbf{56.26}          & 47.24                   \\ \hline
\multicolumn{6}{|c|}{POS}\\
\hline
\multicolumn{1}{|l|}{} & \multicolumn{1}{c|}{EN} & \multicolumn{1}{c|}{DE} & \multicolumn{1}{c|}{AR} & \multicolumn{1}{c|}{HE} & \multicolumn{1}{c|}{ZH} \\ \hline
FL                     & 81.97                   & 88.82                   & 74.07                   & 75.62                   & 61.31                   \\ \hline
MD                     & 82.99                   & 89.19                   & 74.65                   & 77.00                   & 61.74          \\ \hline
SD                     & 82.62                   & 88.74                   & 74.41          & 77.30                   & 61.29                   \\ \hline
EC                     & \textbf{83.49}          & \textbf{89.20}          & \textbf{75.86}                   & \textbf{78.04}          & \textbf{62.33}                   \\ \hline
\end{tabular}
\caption{Cross-lingual NER (upper) and POS (lower) evaluation results with multiple source languages. FL indicates unpruning. MD, SD, and EC are the three heuristics we examine.}
\label{tbl:results-ner-multi-source}
\end{table}

\begin{figure*}[h!]
\centering
\begin{subfigure}{.24\linewidth}
    \centering
    \includegraphics[width=1\linewidth,height=1\linewidth]{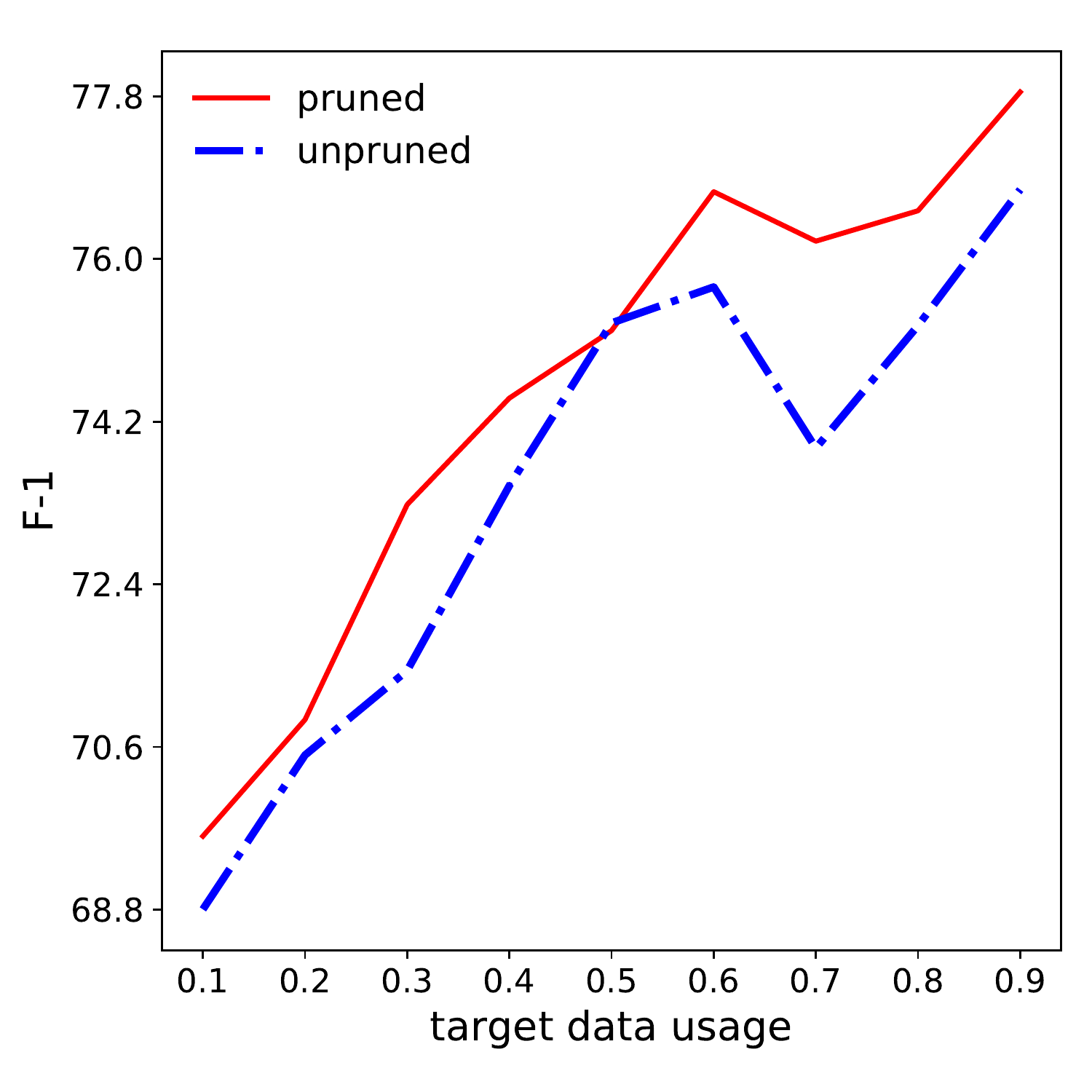}
    \caption{EN-DE}\label{fig:discussion-target-training-size-image11}
\end{subfigure}
    \hfill
\begin{subfigure}{.24\linewidth}
    \centering
    \includegraphics[width=1\linewidth,height=1\linewidth]{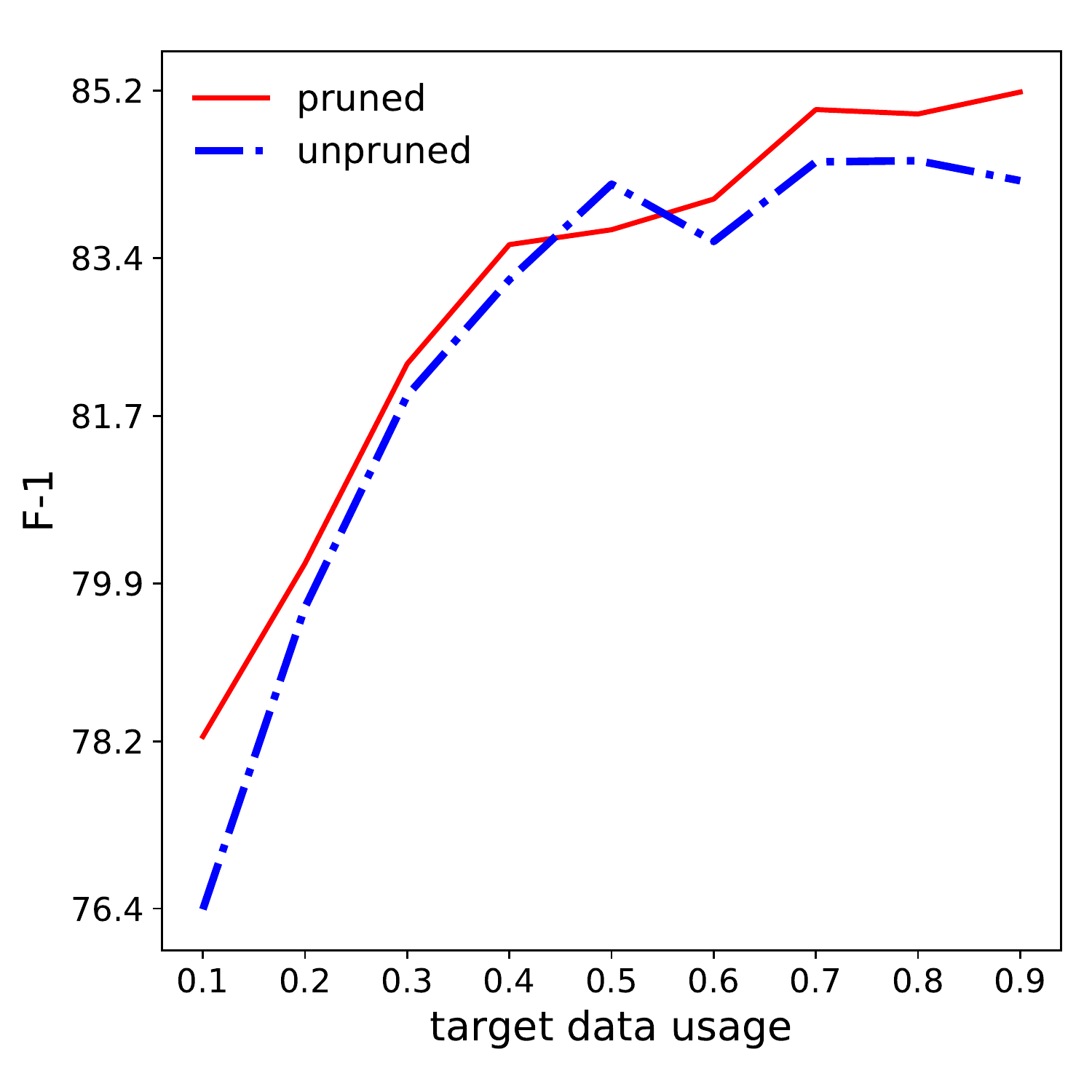}
    \caption{EN-NL}\label{fig:discussion-target-training-size-image12}
\end{subfigure}
   \hfill
\begin{subfigure}{.24\linewidth}
    \centering
    \includegraphics[width=1\linewidth,height=1\linewidth]{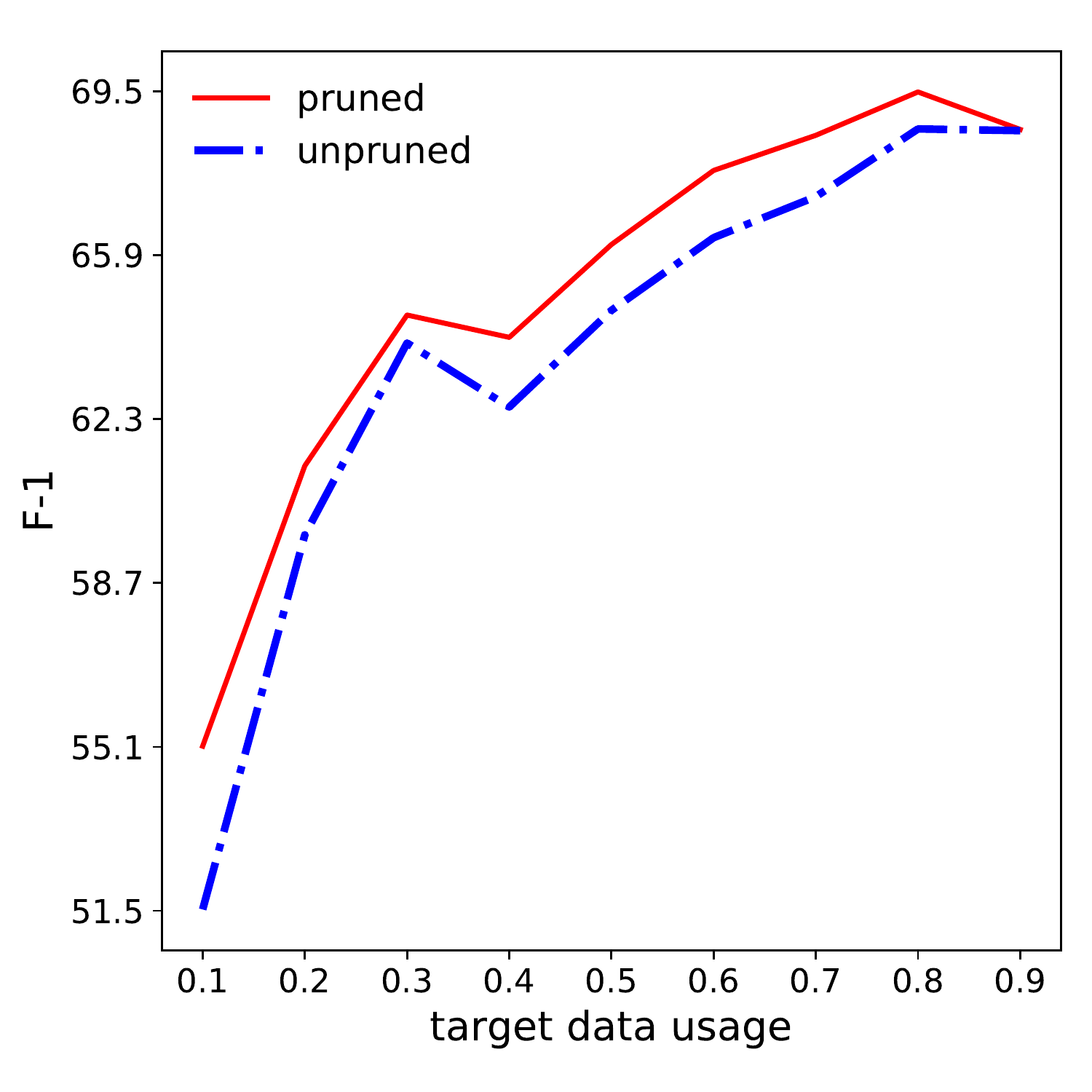}
    \caption{EN-AR}\label{fig:discussion-target-training-size-image13}
\end{subfigure}
   \hfill
\begin{subfigure}{.24\linewidth}
    \centering
    \includegraphics[width=1\linewidth,height=1\linewidth]{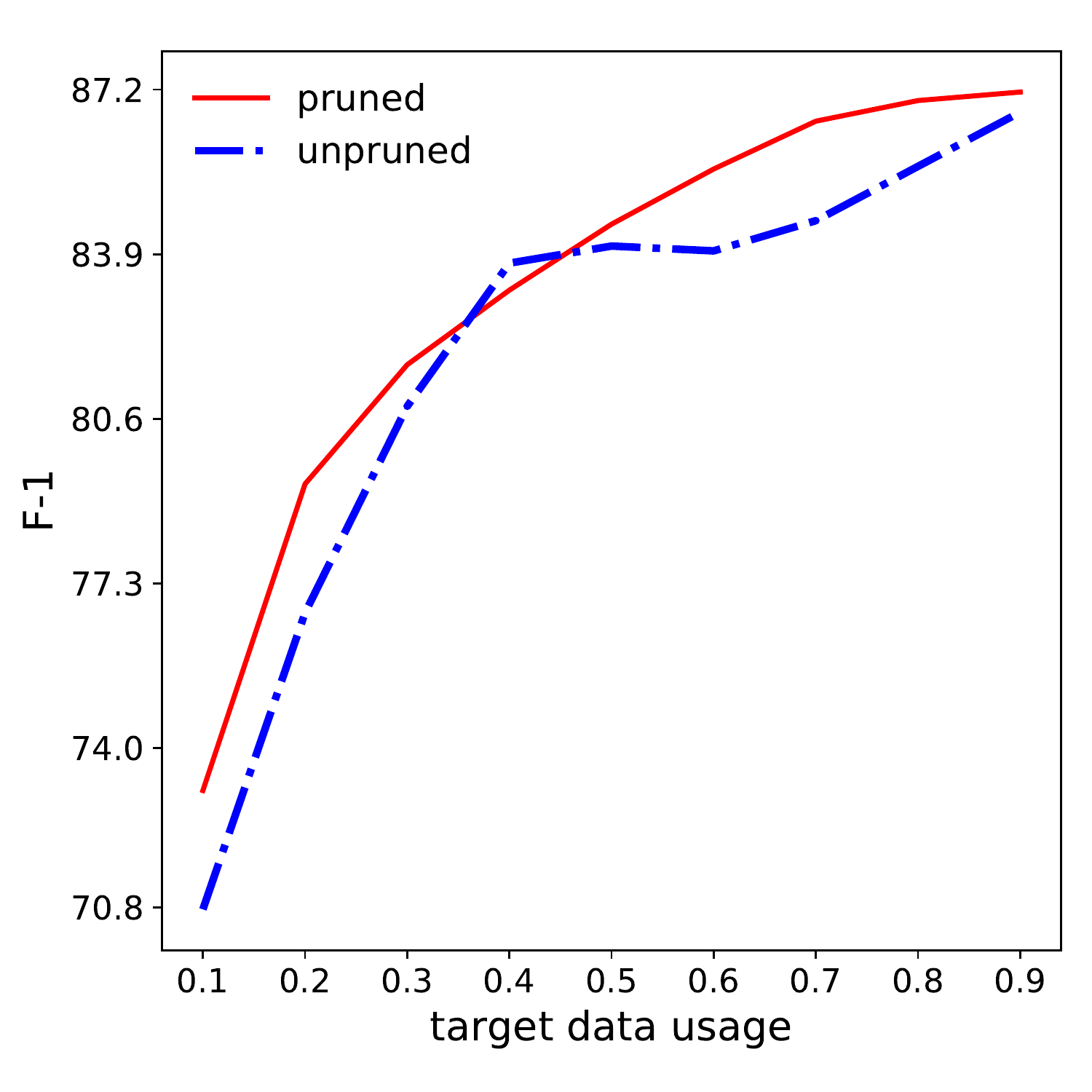}
    \caption{EN-HE}\label{fig:discussion-target-training-size-image14}
\end{subfigure}

\begin{subfigure}{.24\linewidth}
    \centering
    \includegraphics[width=1\linewidth,height=1\linewidth]{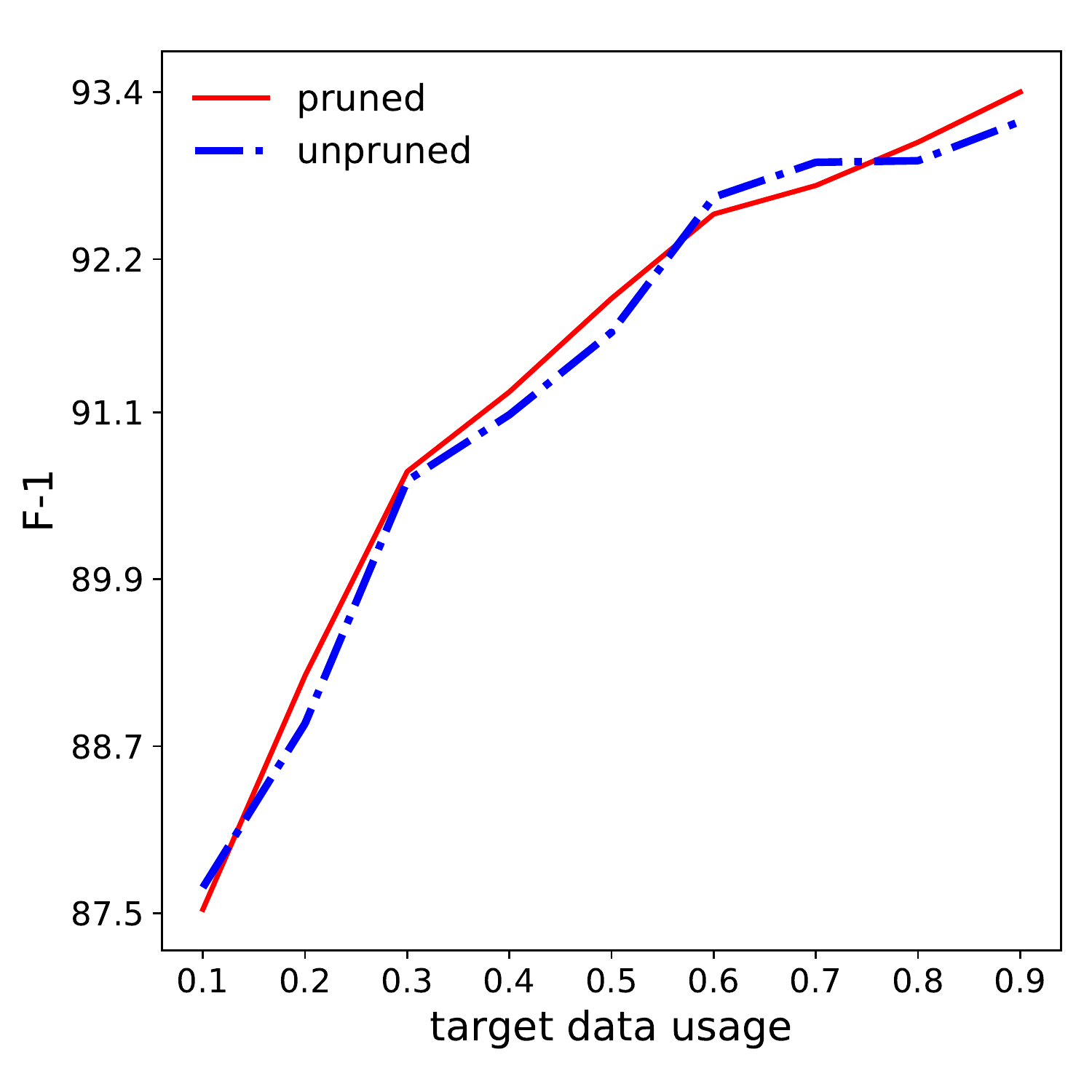}
    \caption{EN-ZH}\label{fig:discussion-target-training-size-image21}
\end{subfigure}
    \hfill
\begin{subfigure}{.24\linewidth}
    \centering
    \includegraphics[width=1\linewidth,height=1\linewidth]{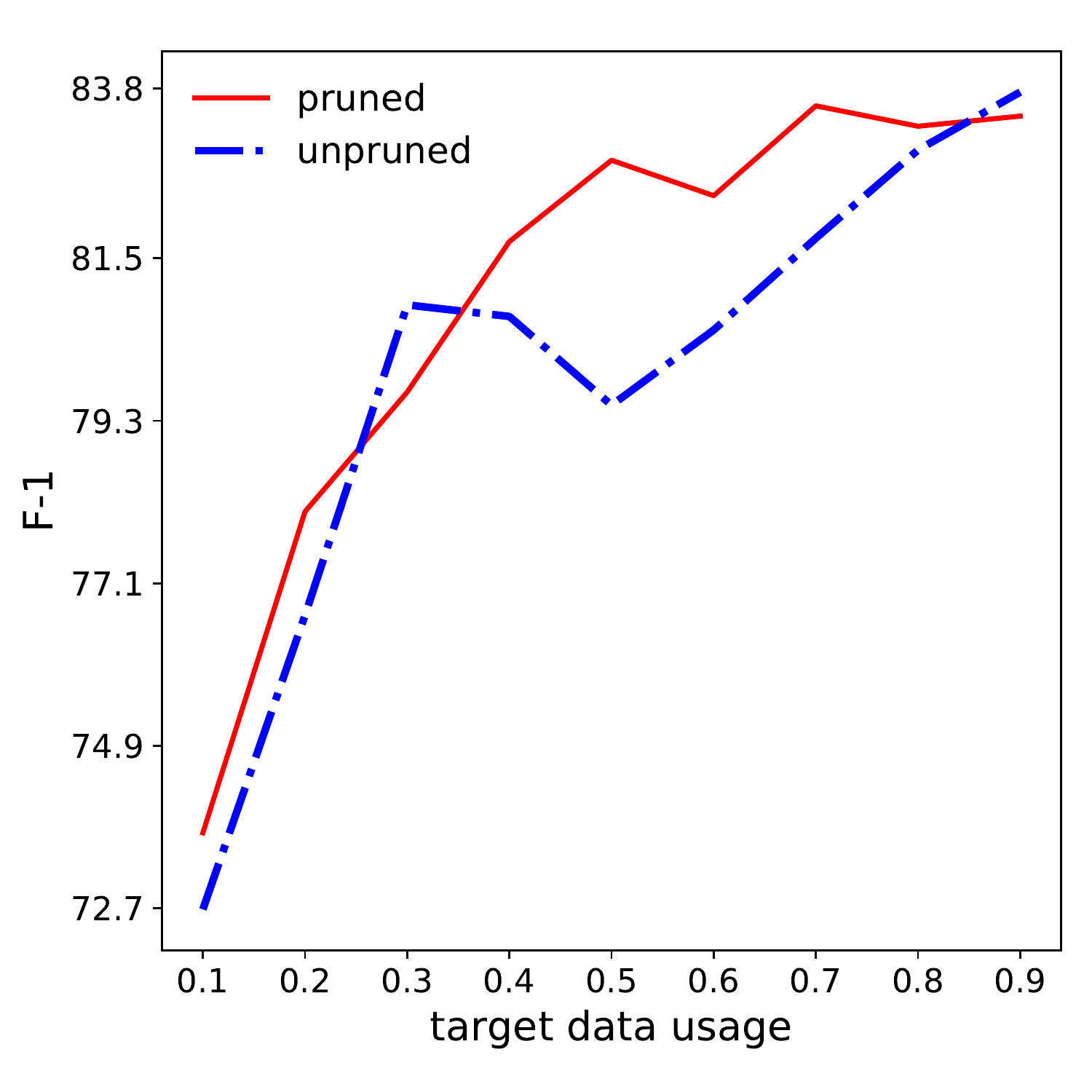}
    \caption{EN-JA}\label{fig:discussion-target-training-size-image22}
\end{subfigure}
   \hfill
\begin{subfigure}{.24\linewidth}
    \centering
    \includegraphics[width=1\linewidth,height=1\linewidth]{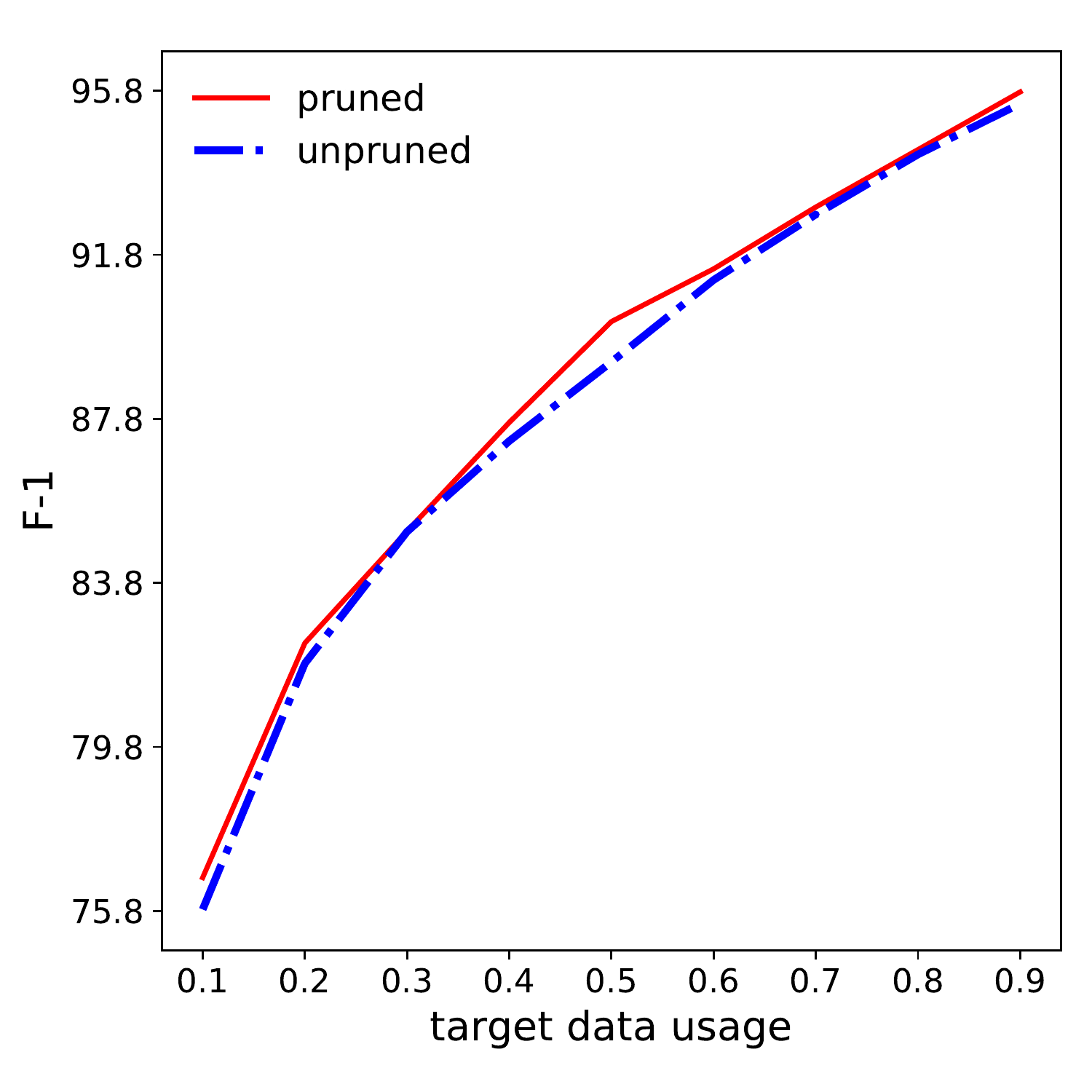}
    \caption{EN-FA}\label{fig:discussion-target-training-size-image24}
\end{subfigure}
\caption{F-1 scores of mBERT on multi-lingual NER with 10\% - 90\% target language training data usage. Dashed blue lines indicate scores without head pruning and solid red lines show scores with head pruning.}
\label{fig:discussion-target-training-size-ner}
\end{figure*}
\begin{figure*}[th]
\centering
\begin{subfigure}{.24\linewidth}
    \centering
    \includegraphics[width=1\linewidth,height=1\linewidth]{pics/en_de_train_rate.pdf}
    \caption{EN-DE}\label{fig:discussion-target-training-size-image11}
\end{subfigure}
    \hfill
\begin{subfigure}{.24\linewidth}
    \centering
    \includegraphics[width=1\linewidth,height=1\linewidth]{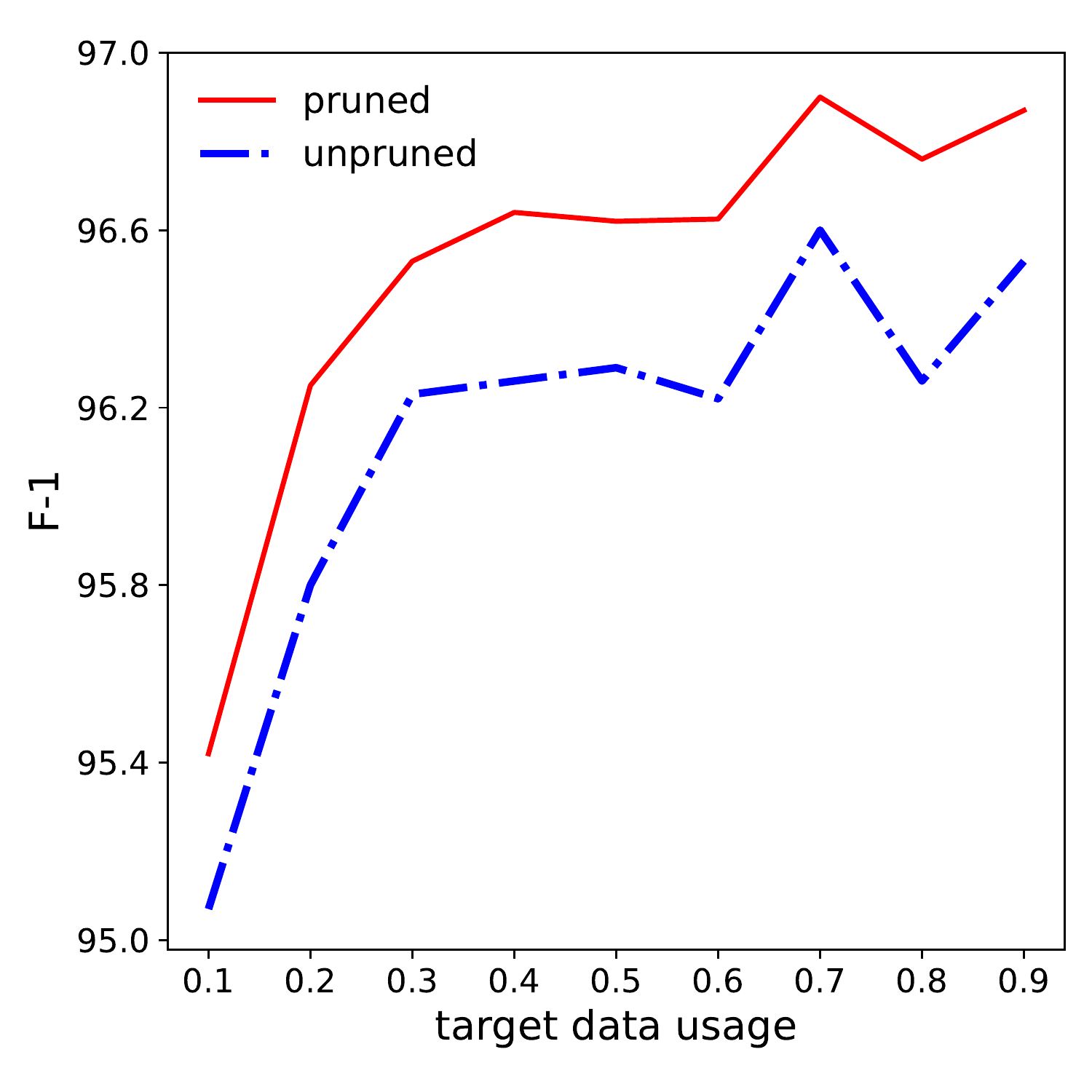}
    \caption{EN-NL}\label{fig:discussion-target-training-size-image12}
\end{subfigure}
   \hfill
\begin{subfigure}{.24\linewidth}
    \centering
    \includegraphics[width=1\linewidth,height=1\linewidth]{pics/en_ar_train_rate.pdf}
    \caption{EN-AR}\label{fig:discussion-target-training-size-image13}
\end{subfigure}
   \hfill
\begin{subfigure}{.24\linewidth}
    \centering
    \includegraphics[width=1\linewidth,height=1\linewidth]{pics/en_he_train_rate.pdf}
    \caption{EN-HE}\label{fig:discussion-target-training-size-image14}
\end{subfigure}

\begin{subfigure}{.24\linewidth}
    \centering
    \includegraphics[width=1\linewidth,height=1\linewidth]{pics/en_cn_train_rate.pdf}
    \caption{EN-ZH}\label{fig:discussion-target-training-size-image21}
\end{subfigure}
    \hfill
\begin{subfigure}{.24\linewidth}
    \centering
    \includegraphics[width=1\linewidth,height=1\linewidth]{pics/en_ja_train_rate.pdf}
    \caption{EN-JA}\label{fig:discussion-target-training-size-image22}
\end{subfigure}
   \hfill
\begin{subfigure}{.24\linewidth}
    \centering
    \includegraphics[width=1\linewidth,height=1\linewidth]{pics/en_fa_train_rate.pdf}
    \caption{EN-FA}\label{fig:discussion-target-training-size-image23}
\end{subfigure}
   \hfill
\begin{subfigure}{.24\linewidth}
    \centering
    \includegraphics[width=1\linewidth,height=1\linewidth]{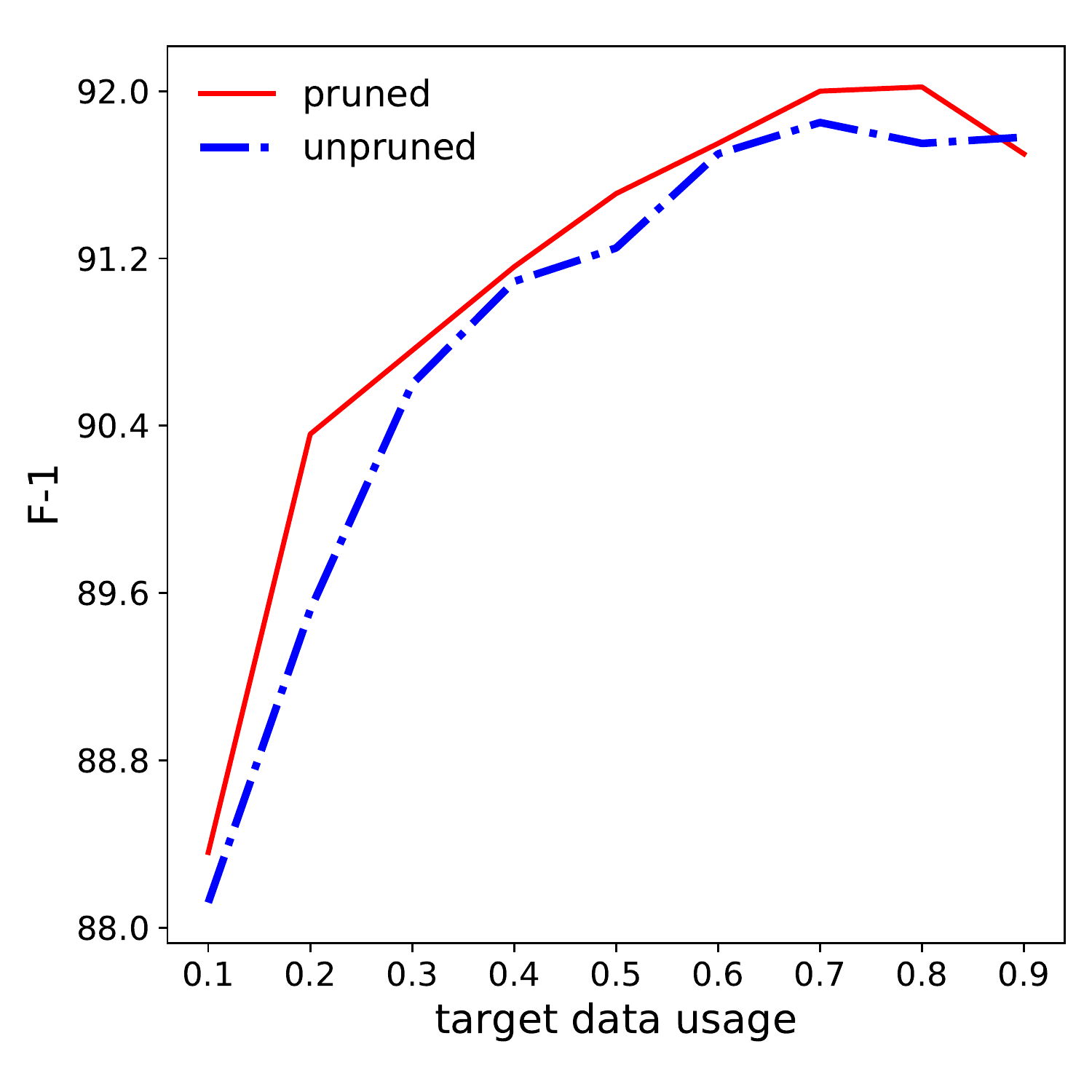}
    \caption{EN-UR}\label{fig:discussion-target-training-size-image24}
\end{subfigure}
\caption{F-1 scores of mBERT on multi-lingual the POS task with 10\% - 90\% target language training data usage. Dashed blue lines indicate scores without head pruning and solid red lines show scores with head pruning.}
\label{fig:discussion-target-training-size-pos}
\end{figure*}

Since the head ranking matrices are not identical across languages, we design three heuristics to rank the heads in the multi-source experiments.
The first method merges the head ranking matrices of all the source languages into one matrix and re-generates the rankings.
The second method ranks the attention heads after summing up the head ranking matrices.
We also examine the efficacy of pruning heads based on the head rankings from a single language.
For this heuristic, we run experiments using the head ranking matrix from each language and report the highest score. We refer to the three heuristics as MD, SD, and EC, respectively.

Table \ref{tbl:results-ner-multi-source} displays the results.
We note that in the NER evaluations, the performance of mBERT on all the languages but ZH are higher than those in the single-source experiments.
This supports our hypothesis that supervision from languages in the same family as the target language helps improve model performance.
Different from NER, the evaluation results on POS are not much higher than the single-source evaluation scores, implying that syntactic features are more consistent across languages than appearances of named entities. 
However, it is consistent on both tasks that pruning attention heads brings performance boosts to all the multi-source experiments.
While the EC heuristic provides the largest improvement margin in 3 out of 5 experiments, it requires a lot more trial experiments.
MD and SD perform comparably well in most cases so they are also promising heuristics for ranking attention heads under the multi-source setting.
The results support that pruning attention heads is beneficial to Transformer-based models in cross-lingual tasks even if the training dataset is already large and diverse in languages.

\subsection{Extension to Resource-poor Languages}

While the languages we use in the main experiments are not truly resource-poor, we examine our findings when training sets in the target languages are smaller.
We design experiments under the multi-lingual setting with subsampled training datasets in target languages.
Specifically, we randomly divide the training set of each target language into 10 disjoint subsets and compare model performance, with and without head pruning, using 1 to 9 subsets. We do not use 0 or 10 subsets since they correspond to cross-lingual and fully multi-lingual settings, respectively.
We run the evaluations on NER and POS tasks.
These datasets vary greatly in size, allowing us to validate our findings on target-language datasets with as few as 80 training examples.
The UR NER dataset is excluded from this case study since its training set is overly large.
We note that the score differences with and without head pruning are, in the main experiments, consistent for all the choices of models and source languages.
Thus, we only display the mBERT performance with EN as the source language on NER in Figure \ref{fig:discussion-target-training-size-ner} and that on POS in Figure \ref{fig:discussion-target-training-size-pos}.

The evaluation results are consistent with those in our main experiments, where the model with up to 12 attention heads pruned generally outperforms the full mBERT model.
This further supports our hypothesis that pruning lower-ranked attention heads has positive effects on the performance of Transformer-based models in truly resource-scarce languages.
It is also worth noting that pruning attention heads often causes the mBERT model to reach peak evaluation scores with less training data in the target language.
For example, in the EN-JA NER experiments, the full model achieves the highest F-1 score when all the 800 training instances in the JA dataset are used while the model with heads pruned achieves a comparable score with 20\% less data. This suggests that pruning attention heads makes deep Transformer-based models easier to train with less training data and thus more applicable to truly resource-poor languages.

\section{Conclusion and Future Work}
This paper studied the contributions of attention heads in Transformer-based models.
Past research has shown that in mono-lingual tasks, pruning a large number of attention heads can achieve comparable or higher performance than the full models.
However, we were the first to extend these findings to cross-lingual and multi-lingual sequence labeling tasks.
Using a gradient-based method, we identified the heads to prune and showed that pruning attention heads generally has positive effects on mBERT and XLM-R performances.
Additional case studies empirically demonstrated the validity of our findings and showed further extensibility of them to a wider range of task settings.
In addition to better understanding of Transformer-based models under cross- and multi-lingual settings, our findings can be applied to existing models to achieve better performance with reduced training data and resource consumption.
Future work could include improving model interpretability in other cross-lingual and multi-lingual tasks, e.g. XNLI \cite{xnli} and other passage-level classification tasks.

\bibliography{anthology,acl2020}
\bibliographystyle{acl_natbib}

\end{document}